%% file: neurips_2026.tex
\documentclass{article}

\usepackage[main, final]{neurips_2026}

\usepackage[utf8]{inputenc}
\usepackage{hyperref}
\usepackage{url}
\usepackage{booktabs}
\usepackage{amsfonts}
\usepackage{nicefrac}
\usepackage{amsmath,amsthm}
\usepackage{paralist,bm}
\usepackage{algorithm,algorithmic}
\usepackage{graphicx}
\usepackage{makecell}
\usepackage{subfigure}
\usepackage{pifont}
\usepackage[table]{xcolor}
\usepackage{threeparttable}
\theoremstyle{plain}
\newtheorem{theorem}{Theorem}[section]

\newtheorem{lemma}[theorem]{Lemma}

\theoremstyle{definition}
\newtheorem{definition}[theorem]{Definition}
\newtheorem{assumption}[theorem]{Assumption}
\theoremstyle{remark}

\def \E {\mathbb{E}}
\def \g {\mathbf{g}}
\def \Lb {\mathbf{L}}
\def \w {\mathbf{w}}

\def \z {\mathbf{z}}

\def \u {\mathbf{u}}
\def \x {\mathbf{x}}
\def \R {\mathbb{R}}

\def \P {\mathcal{P}}
\def \A {\mathcal{A}}
\def \p {\mathbf{p}}
\def \q {\mathbf{q}}
\def \I {\mathcal{I}}
\def \Fb {\mathbf{F}}
\def \s {\mathbf{s}}

\def \W {\mathcal{W}}
\def \qb {\bar{\q}}
\def \vv {\mathbf{v}}
\def \wb {\bar{\w}}
\def \wt {\widetilde{\w}}

\def \N {\mathbb{N}^+}

\def \ind {\mathbb{I}}

\def \Regqp {\mathrm{Regret}_{q,t}^{\prime}}
\def \Regq {\mathrm{Regret}_{q,t}}
\def \Regw {\mathrm{Regret}_{w,t}}

\newcommand\inner[2]{\langle #1, #2 \rangle}
\def \hats {\hat{s}}
\def \hatsb {\hat{\s}}
\def \tildes {\tilde{s}}
\def \tildesb {\tilde{\s}}
\newcommand \indicator[1]{\mathbb{I}[#1]}

\DeclareMathOperator*{\argmin}{argmin}
\DeclareMathOperator*{\argmax}{argmax}

\title{Group Distributionally Robust Optimization with Flexible Sample Queries}

\author{
Haomin Bai$^{1,2}$ \quad
Dingzhi Yu$^{1,2}$ \quad
Shuai Li$^{3,4}$ \quad
Haipeng Luo$^5$ \quad
Lijun Zhang$^{1,2}$\thanks{Lijun Zhang is the corresponding author.}\\
$^1$\,State Key Laboratory of Novel Software Technology, Nanjing University, Nanjing, China\\
$^2$\,School of Artificial Intelligence, Nanjing University, Nanjing, China\\
$^3$\,The University of Hong Kong, Hong Kong, China \\
$^4$\,Shanghai Innovation Institute, Shanghai, China\\
$^5$\,Thomas Lord Department of Computer Science, University of Southern California, USA\\
\texttt{\{baihm,yudz,zhanglj\}@lamda.nju.edu.cn}\\
\texttt{shuaili@hku.hk \quad haipengl@usc.edu}
}

\begin{document}

\maketitle

\begin{abstract}
Group distributionally robust optimization (GDRO) aims to learn models that perform well across $m$ distributions simultaneously.
Existing methods either (i) query $m$ samples per round, which imposes strict implementation requirements, or (ii) query only a single sample per round, which requires many rounds to converge and therefore leads to long runtime.
To avoid these limitations, we introduce a novel setting called \emph{flexible sample queries} (FSQ), which allows the number of samples that can be queried to vary across rounds.
Under the FSQ setting, we cast GDRO as a two-player game: one performs non-oblivious online convex optimization, while the other tackles a non-oblivious prediction with limited advice (PLA) problem. 
In this game, the first player updates its decision using follow-the-regularized-leader with stochastic gradients.
For the second player, we develop a new PLA method equipped with an adaptive exploration strategy that can leverage an arbitrary per-round sample size. 
We then establish the \emph{first} high-probability regret bound for non-oblivious PLA.
By integrating these components, we develop an anytime GDRO algorithm that supports FSQ, and derive a high-probability optimization error bound of $O\left(\frac{1}{t}\sqrt{\sum_{j=1}^t \frac{m}{r_j}\log m}\right)$, where $r_j$ is the sample size at round $j$. 
This result demonstrates that the optimization error decreases as the per-round sample sizes increase, and implies the same near-optimal sample complexity of $O(m\log (m)/\epsilon^2)$ for any fixed sample size $r\in[m]$.
\end{abstract}

\section{Introduction}
Traditional machine learning methods typically train models by minimizing the empirical risk over random samples drawn from an unknown distribution \citep{Nature_Statistical}. 
However, when the test data distribution deviates from the training data distribution, models often suffer significant performance degradation \citep{JMLR:v8:sugiyama07a,10.1145/2523813,distribution_shift}. 
Distributionally robust optimization (DRO) is a promising approach to mitigating the effect of distribution shifts by minimizing the worst-case risk over an uncertainty set \citep{delage2010distributionally, DRO, Rahimian_2022}. 

In this paper, we focus on a special case of DRO, known as group DRO (GDRO) \citep{GDRO-oren2019,Gouop_DRO_ICLR_2020}, where the uncertainty set is defined as a finite collection of distributions. GDRO can be mathematically formulated as a minimax stochastic optimization problem
\begin{equation} \label{eqn:group:dro}
	\min_{\w \in \W}   \max_{i\in[m]}  \ \left\{ R_i(\w)=\E_{\z\sim\P_i}\left[\ell(\w;\z)\right]\right\},
\end{equation}
where $\{\P_i\}_{i\in[m]}$ denotes a set of $m$ distributions, $\z$ is a random sample, $\w$ corresponds to the model, $\W$ represents the hypothesis class, and $\ell(\cdot;\cdot)$ is the loss function.
Assuming that each risk function $R_i(\w)$ is convex for all $i\in[m]$, the optimization problem in \eqref{eqn:group:dro} can be cast as a stochastic convex-concave saddle-point problem \citep{nemirovski-2008-robust}
\begin{equation} \label{eqn:GDRO:convex:concave}
	\min_{\w \in \W} \max_{\q \in \Delta_m}  \   \left\{\phi(\w,\q)= \sum_{i=1}^m q_i R_i(\w) \right\},
\end{equation}
where $\Delta_m=\{\q \in \R^m| \q \geq\mathbf{0}_m, \sum_{i=1}^m q_i=1\}$ denotes the ($m{-}1$)-dimensional simplex.

Recent work has proposed various algorithms for solving GDRO in~\eqref{eqn:group:dro}/\eqref{eqn:GDRO:convex:concave} and established a nearly optimal sample complexity of $O(m \log(m)/\epsilon^2)$, where $\epsilon$ denotes the error tolerance for~\eqref{eqn:group:dro}. 
These methods assume sampling access to any of the $m$ distributions, but restrict the per-round sample size to either $m$~\citep{nemirovski-2008-robust} or $1$~\citep{Online:Multiple:Distribution,DRO:Online:Game,NEURIPS2022,zhang2023-SA-GDRO}.
In many applications, however, the per-round sampling capacity is constrained by time-varying external resources, such as communication bandwidth in mobile-edge federated learning~\citep{2019_FedCS}, human review capacity in content moderation~\citep{2023_FairM_KDD}, and expert annotation capacity in medical data collection~\citep{2024_Survey_Medical}. In these settings, the learner can still decide which groups to query under the available sampling capacity. We provide a detailed discussion of these examples in Appendix~\ref{sec:applications}.

To capture such scenarios, we introduce a new setting termed \emph{flexible sample queries} (FSQ), where at the beginning of each round $t$, the environment reveals a potentially time-varying sample size $r_t\in[m]$. After observing $r_t$, the learner decides which groups to query.
Existing GDRO algorithms are not satisfactory in the FSQ setting for the following reasons:
\begin{compactenum}
    \item
    Methods whose updates require $m$ samples cannot proceed when the available sample size is less than $m$. They must wait until enough samples are collected to form an update, which introduces latency.
    \item
    Methods that process one sample per update need to handle the $r_t$ samples sequentially, which substantially increases the number of iteration rounds and leads to a long runtime. 
\end{compactenum}
Overall, existing methods are tailored to extreme cases and cannot adapt to varying sample sizes.

In this paper, we study GDRO under the FSQ setting and, following prior work, formulate it as a two-player game~\citep{Hedge_1997}, where the minimizer is the $w$-player and the maximizer is the $q$-player. 
For the $w$-player, we model its task as ``non-oblivious online convex optimization (OCO) with stochastic gradients''~\citep{zhang2023-SA-GDRO}, and update its decisions using follow-the-regularized-leader (FTRL)~\citep{orabona2023}. 
By constructing an unbiased stochastic gradient from a single queried sample, FTRL yields guarantees independent of the per-round sample size.
Our main technical contribution is to design a $q$-player algorithm that can operate with a variable number of samples per round. To this end, we model the $q$-player’s task as ``non-oblivious prediction with limited advice (PLA)''. However, existing PLA methods are not applicable to our setting, as they either do not handle non-oblivious losses~\citep{pmlr-v32-seldin14} or cannot accommodate varying sample sizes~\citep{Twolayers-ACM-2018}. 
We therefore propose \textsc{NPLA-Exp3}, an Exp3-style algorithm for non-oblivious PLA, which employs an adaptive exploration strategy based on the sample size \(r_t\).
When $r_t \geq 2$, \textsc{NPLA-Exp3} performs \emph{active} exploration by allocating $1$ sample to exploitation and the remaining $r_t - 1$ to exploration, yielding an unbiased loss estimator as in~\citet{pmlr-v32-seldin14}. 
When $r_t = 1$, \textsc{NPLA-Exp3} resorts to \emph{implicit} exploration by adding an offset to the selection probability, resulting in the biased Implicit-eXploration (IX) estimator~\citep{NIPS14:IX-estimator}.
The loss estimators are accumulated over rounds to compute $\q_t$ via FTRL with negative entropy.
Based on this design, we establish the \emph{first} high-probability regret bound $O\left(\sqrt{\sum_{j=1}^t \frac{m}{r_j} \log m}\right)$ for non-oblivious PLA.

Leveraging the methods for the two players, we develop \textsc{Flex-GDRO} and establish a high-probability optimization error bound that depends on the per-round sample size $r_t$. In addition, \textsc{Flex-GDRO} runs in an \emph{anytime} manner, meaning that it does not require prior knowledge of the number of iterations and can return a solution at any round~\citep{pmlr-v97-cutkosky19a,zhang2026-GDRO-TPAMI}.
Our contributions are summarized as follows:
\begin{compactenum}
	\item We introduce the FSQ setting and develop \textsc{Flex-GDRO}, the first GDRO algorithm that can process an arbitrary number of samples per round. Then, we establish a high-probability optimization error bound of $O\left(\frac{1}{t}\sqrt{\sum_{j=1}^t \frac{m}{r_j}\log m}\right)$.
    \item  When the per-round sample size is fixed, i.e., $r_t \equiv r$, \textsc{Flex-GDRO} improves upon existing methods in three respects:
    (i) it possesses the anytime ability while matching the near-optimal sample complexity $O(m \log(m)/\epsilon^2)$ achieved by non-anytime algorithms~\citep{nemirovski-2008-robust,zhang2023-SA-GDRO};
    (ii) it improves the anytime optimization error bound from $O(\log(t)\sqrt{m \log(m)/(r t)})$~\citep{zhang2026-GDRO-TPAMI} to $O(\sqrt{m \log(m)/(r t)})$; and
    (iii) it supports any $r\in[m]$, rather than $r\in\{1,m\}$.
    \item As a by-product, we propose a novel non-oblivious PLA algorithm, \textsc{NPLA-Exp3}, and establish the first high-probability regret bound $O\left(\sqrt{\sum_{j=1}^t \frac{m}{r_j}\log m}\right)$, which is of the same order as the expected regret bound against an oblivious adversary~\citep{pmlr-v32-seldin14}.
\end{compactenum}

\paragraph{Related Work.} 
We briefly review related work on GDRO and PLA, deferring a more detailed discussion to Appendix~\ref{app:related_work}.
GDRO is motivated by applications affected by distribution shift~\citep{2019_Agnostic_FL,GDRO-oren2019,Gouop_DRO_ICLR_2020}.
Existing methods fall into two fixed sampling regimes: querying one sample from each of the $m$ distributions per round~\citep{nemirovski-2008-robust}, or querying one sample from a single distribution per round~\citep{NEURIPS2022,Online:Multiple:Distribution,DRO:Online:Game,zhang2023-SA-GDRO}.
Existing anytime variants are also restricted to the two extremes $r\in\{1,m\}$~\citep{zhang2026-GDRO-TPAMI}.
The learning problem faced by the $q$-player is related to PLA~\citep{pmlr-v32-seldin14}, which lies between multi-armed bandits (MAB)~\citep{Robbins-1952,Auer02} and prediction with expert advice~\citep{Hedge_1997}. In PLA, the player queries the advice of $r_t\in[m]$ experts in round $t$, and the loss is evaluated on a subset of the selected experts~\citep{pmlr-v32-seldin14,pmlr-v201-saad23a}.
Existing PLA algorithms are inapplicable to our setting, which requires minimizing regret against a non-oblivious adversary with varying sample sizes, because prior work assumes stochastic adversaries~\citep{NEURIPS2021_c688defd}, oblivious adversaries~\citep{pmlr-v32-seldin14,pmlr-v201-saad23a},  fixed sample sizes~\citep{NEURIPS2018_253f7b5d,Twolayers-ACM-2018} or focuses on pseudo-regret objectives~\citep{pmlr-v35-kale14a,Amin_Kale_Tesauro_Turaga_2015}.

\section{Preliminaries and Two-Player Game Formulation}  \label{sec:preliminary}
We begin by equipping the domain $\W$ with a function $\nu_w(\cdot)$ that is $1$-strongly convex with respect to the norm $\|\cdot\|_w$. 
Similarly, the simplex $\Delta_m$ is equipped with the negative entropy function $\nu_q(\q)=\sum_{i=1}^m$ $ q_i \ln q_i$, which is $1$-strongly convex with respect to the $\ell_1$-norm $\|\cdot\|_1$.
Next, we make the following standard assumptions~\citep{nemirovski-2008-robust,Online:Multiple:Distribution}.
\begin{assumption}\label{ass:gen_domain_w} 
	$\W$ is convex and its diameter measured by a function $\nu_w(\cdot)$ is bounded by $D$, i.e.,
	\begin{equation} \label{eqn:ass:domain:W}
		\max_{\w \in \W} \nu_w(\w) -\min_{\w \in \W} \nu_w(\w)  \leq D^2.
	\end{equation}
\end{assumption}
It is immediate that $\Delta_m$ is convex, and its diameter measured by $\nu_q(\cdot)$ is bounded by $\sqrt{\ln m}$.
\begin{assumption}\label{ass:gen_gradient_bounded}
	For all $i\in[m]$, we assume
	\begin{equation} \label{eqn:ass:gradient_bound}
		\left\|\nabla \ell\left(\w;\z\right)\right\|_{w,*}\leq G \quad  \forall\w\in\W, \ \z\sim\P_{i},
	\end{equation}
	where $\|\cdot\|_{w,*}$ denotes the dual norm of $\|\cdot\|_{w}$.
\end{assumption}
\begin{assumption}\label{ass:gen_loss_bounded} 
	For all $i\in[m]$, we assume
	\begin{equation} \label{eqn:ass:value_bound}
		0\leq \ell\left(\w;\z\right) \leq 1 \quad  \forall\w\in\W, \ \z\sim\P_{i}.
	\end{equation}
\end{assumption}
\begin{assumption}\label{ass:gen_loss_convex}
	For all $i\in[m]$, the risk function $R_i(\w)=\E_{\z \sim \P_i}[\ell(\w;\z)]$ is convex.
\end{assumption}
For any approximate solution $(\wb, \qb)$ of problem~\eqref{eqn:GDRO:convex:concave}, its optimization error is defined as 
$$\epsilon_{\phi}(\wb, \qb) = \max_{\q \in \Delta_m} \phi(\wb, \q) - \min_{\w \in \W} \phi(\w, \qb),$$ 
which controls the optimality of $\wb$ for the original GDRO problem~\eqref{eqn:group:dro} \citep{zhang2023-SA-GDRO}.

\paragraph{Error Decomposition via a Two-Player Game.}
We cast the problem~\eqref{eqn:GDRO:convex:concave} as a two-player game~\citep{Hedge_1997}: in each round $t$, the $w$-player selects $\w_t\in\W$ to minimize the convex function $\phi(\w,\q_t)= \sum_{i=1}^m q_{t,i} R_i(\w)$, and the $q$-player selects $\q_t\in\Delta_m$ to maximize the linear function $\phi(\w_t,\q)= \sum_{i=1}^m q_i R_i(\w_t)$.
Based on iterative decisions, we output the averaged solutions $(\bar{\w}_t, \bar{\q}_t)$ at each iteration $t$, defined as
\begin{equation}\label{eqn:average_output}
	\wb_t=\frac{1}{t} \sum_{j=1}^t \w_j,~\text{and}~\qb_t=\frac{1}{t} \sum_{j=1}^t \q_j.
\end{equation}
Under Assumption~\ref{ass:gen_loss_convex} and by Jensen’s inequality, the optimization error can be decomposed into the sum of two average regrets
\begin{equation}  \label{eqn:decom:1}
	\begin{split}
	&\epsilon_{\phi}(\wb_t, \qb_t)\\
     \leq & \frac{1}{t}\underbrace{\left(\sum_{j=1}^t \phi(\w_j,\q_j) - \min_{\w\in \W}  \sum_{j=1}^t \phi(\w,\q_j) \right)}_{\Regw} + \frac{1}{t}\underbrace{\left(\max_{\q\in \Delta_{m}}  \sum_{j=1}^t \phi(\w_j,\q) - \sum_{j=1}^t \phi(\w_j,\q_j) \right)}_{\Regq},
	\end{split}
\end{equation}
where $\Regw$ and $\Regq$ are the regrets of the $w$-player and the $q$-player, respectively.

\paragraph{Learning Problems for the Two Players.} 
In the game formulation, both players face \emph{non-oblivious} online learning problems, since the objective functions encountered by each player depend on their past decisions.
Moreover, the players cannot observe the exact values of risks $R_i(\w)$ or gradients $\nabla R_i(\w)$. Instead, they only have access to stochastic estimates using randomly queried samples.
Building on prior analyses~\citep{zhang2023-SA-GDRO}, we observe that for the $w$-player, $1$ sample drawn according to $\q_t$ suffices to form an unbiased stochastic gradient of $\phi(\w, \q_t)$ with low norm under Assumption~\ref{ass:gen_gradient_bounded}. Therefore, we model the problem faced by the $w$-player as ``non-oblivious OCO with stochastic gradients'', whose performance does not depend on the per-round sample size $r_t$.
In contrast, for the $q$-player, $r_t$ has a direct impact on $\Regq$.
To make this dependence explicit, we decouple the stochastic nature (arising from stochastic observation) from the adversarial nature (non-obliviousness in the game) as follows, in a manner similar to~\citet{Online:Multiple:Distribution}.

We first transform the maximization problem over $\{\phi(\w_j,\q)=\sum_{i=1}^m q_i R_i(\w_j)\}_{j\in[t]}$ into an equivalent minimization problem over $\{\sum_{i=1}^m q_i (1 - R_i(\w_j))\}_{j\in[t]}$ under Assumption~\ref{ass:gen_loss_bounded}.
Next, we define the surrogate loss
\begin{equation}  \label{def:hats:main_part}
	\hats_{t,i}= 1 - \ell(\w_t; \z_t^{(i)})\overset{\eqref{eqn:ass:value_bound}}{\in}  [0,1], \quad \forall i\in[m],
\end{equation}
where each $\z_{t}^{(i)} \sim \P_i$ is a random sample drawn at round $t$. We note that $\hatsb_t$ is introduced for analysis, and only the queried coordinates $\hats_{t,i}$ are observed. For any fixed $t\in\mathbb{N}_+$ and $\delta\in(0,1)$, concentration inequalities~\citep{cesa2006prediction} imply that, with probability at least $1-\delta$, 
\begin{align*}
    \Regq  = \sum_{j=1}^t \inner{\q_j}{\E[\hatsb_j]} - \min_{i\in[m]} \sum_{j=1}^t \E[\hats_{j,i}] 
    \leq \underbrace{\sum_{j=1}^t \inner{\q_j}{\hatsb_j} - \min_{i\in[m]} \sum_{j=1}^t \hats_{j,i}}_{\Regqp} + O\left(\sqrt{t\log \frac{m}{\delta}}\right),
\end{align*}
where the last term  captures the stochastic nature. We then assign the $q$-player to minimize $\Regqp$ with $r_t$ observations per round, which yields a ``non-oblivious PLA'' problem. 

\textbf{Remark:}
One may attempt to cast the $q$-player's problem as a combinatorial semi-bandit~\citep{Semi-bandits-2014,JMLR:v17:15-091}.
However, methods for combinatorial semi-bandits are not applicable because they compete with the best subset of $r$ experts in terms of average loss, which does not control $\Regq$ defined against the best \emph{single} expert.

Combining the above results, the following bound holds with probability at least $1-\delta$ (see Lemma~\ref{lem:General_framework} for a complete statement),
\begin{equation}  \label{eqn:decom:2}
        \epsilon_{\phi}(\wb_t, \qb_t)  \leq\frac{1}{t}\Regw + \frac{1}{t}\Regqp+O\left(\sqrt{\frac{1}{t}\log \frac{m}{\delta}}\right).
\end{equation}
Therefore, it suffices to design no-regret online learning algorithms for each player, and combining their regret bounds yields the overall optimization error.
In particular, the resulting $q$-player algorithm is established for non-oblivious PLA and may be of independent interest beyond GDRO.

\section{\textsc{Flex-GDRO} for GDRO with FSQ}
We introduce \textsc{Flex-GDRO} through three components: (i) the $q$-player strategy $\A_q$, (ii) the $w$-player strategy $\A_w$, and (iii) a meta-scheduler $\A_s$ that interacts with the environment and coordinates the two players.
Since the $w$-player can reuse the samples queried by the $q$-player (see Section~\ref{sec:w-player}), we first describe $\A_q$, followed by $\A_w$, and finally $\A_s$. The sample-size protocol is defined as follows.
\begin{definition}[Sample-size protocol]\label{def:sample_size_protocol}
At the beginning of each round \(t\), the environment reveals a sample size \(r_t \in [m]\) before the learner's round-\(t\) randomization and before fresh samples are drawn. The sample size \(r_t\) may depend on all history before round \(t\) and on exogenous randomness.
\end{definition}

\subsection{\textsc{NPLA-Exp3} for the $q$-player}  \label{sec:q-player}
Recall that the $q$-player needs to solve a non-oblivious PLA problem, to which existing methods are not applicable (see Appendix~\ref{app:related_work} for a detailed discussion).
A key challenge in achieving low regret is to appropriately balance exploitation and exploration, as in MAB~\citep{Bandit:survey}.
To address this challenge, we propose an adaptive exploration strategy based on the per-round sample size $r_t$:
\begin{compactenum}
    \item When $r_t \geq 2$, we use 1 sample for exploitation and the remaining $r_t - 1$ samples for exploration via uniform selection over the other $m-1$ groups. This active exploration strategy, first applied in oblivious PLA~\citep{pmlr-v32-seldin14}, yields an unbiased estimator with low variance.
    \item When $r_t = 1$, active exploration is infeasible because we only observe the loss of the exploited option. In this case, we adopt \emph{implicit} exploration via the IX estimator~\citep{NIPS14:IX-estimator}, which injects a small bias into the loss estimator. 
\end{compactenum}
The two loss estimators induced by these regimes are then combined to update $\q_t$.
We detail the procedure below and provide theoretical guarantees via a novel analysis.

At the start of each round $t$, the $q$-player receives $r_t$ from the meta-scheduler $\A_s$ and then updates $\q_t$ via FTRL as
\begin{equation}\label{eqn:GDRO:qt_update_FTRL:col}
	\q_t=\argmin_{\q\in \Delta_m}\left\{\inner{\eta_{q,t}\Lb_{t-1}}{\q}+\nu_q(\q)\right\},
\end{equation}
where $\Lb_{t-1}\in\R^m$ is the cumulative loss vector initialized as $\Lb_0=\mathbf{0}_m$, and the step size is set as
\begin{equation}\label{eqn:proof:q:stepsize:1}
	\eta_{q,t}=\sqrt{\frac{\ln m}{m\sum_{j=1}^t\frac{1}{r_j}}}.
\end{equation} 
With the regularizer $\nu_q(\q)=\sum_{i=1}^m q_i\ln q_i$, the FTRL update in~\eqref{eqn:GDRO:qt_update_FTRL:col} reduces to the exponential-weights update rule
\begin{equation}\label{eqn:GDRO:qt_update:col}
	q_{t,i}=\frac{\exp(-\eta_{q,t}L_{t-1,i})}{\sum_{k=1}^m\exp(-\eta_{q,t}L_{t-1,k})}, \quad \forall i\in[m].
\end{equation}

We proceed by querying new samples through the following process. The initial step is to select a distribution index $c_t\in[m]$ according to $\q_t$. For rounds where $r_t\geq2$, we additionally select $r_t-1$ distribution indices uniformly at random and without replacement from the remaining $m-1$ elements. These additional indices form the set $\I_t$ and the complete set of selected indices is denoted by $C_t = \{c_t\} \cup \I_t$.
The sampling of $\I_t$ can be implemented using the DepRound algorithm \citep{DepRound} with $O(m)$ time and space complexity, as detailed in Appendix~\ref{appendix:section:DepRound}.
Other sampling procedures that yield the required subset size and marginal inclusion probabilities can also be used.

Next, the set $C_t$ together with the exploited index $c_t$ are sent to the scheduler $\A_s$, which returns the realized losses $\hats_{t,i}$ for all $i\in C_t$.
We then construct an adaptive loss estimator $\tildesb_t$ based on the per-round sample size $r_t$ for each $i\in[m]$ as
\begin{equation}  
	\label{eqn:GDRO:loss_estimator}
	\begin{split}
		\tilde{s}_{t,i} = 
		\begin{cases}
			\displaystyle
			\frac{\hat{s}_{t,i}}{q_{t,i} + \gamma_t} \ind\left[i \in C_t \right],
			& r_t = 1 \\[11pt]
			\displaystyle
			\frac{\hat{s}_{t,i}}{q_{t,i} + (1 - q_{t,i}) \frac{r_t - 1}{m - 1}} \ind\left[i \in C_t \right],
			&  r_t \geq 2
		\end{cases},
	\end{split}
\end{equation}
where $\hats_{t,i}$ is defined in \eqref{def:hats:main_part}, and the IX coefficient $\gamma_t$ is set to
\begin{equation}   \label{eqn:proof:q:stepsize:sep:gammat}
	\gamma_t=\frac{1}{2}\eta_{q,t}.
\end{equation}
As can be seen from~\eqref{eqn:GDRO:loss_estimator}, when $r_t=1$, we adopt the (biased) IX estimator~\citep{NIPS14:IX-estimator}, which replaces the true observation probability $q_{t,i}$ with the IX-adjusted term $q_{t,i}+\gamma_t$. The parameter $\gamma_t$ implements \emph{implicit} exploration and prevents overly aggressive importance weighting when $q_{t,i}$ is small.
When $r_t\ge 2$, we instead use the true observation probability induced by the sampling rule, which yields an unbiased estimator~\citep{pmlr-v32-seldin14}.

At the end of each round $t$, the cumulative loss is updated as
\begin{equation}    \label{eqn:GDRO:cumulative_loss_estimator:col}
	\Lb_{t}=\Lb_{t-1}+\tildesb_{t}.
\end{equation}
Since our algorithm is designed for non-oblivious PLA and adopts an \textsc{Exp3}-style update rule~\citep{Auer02} with adaptive exploration, we term it \textsc{NPLA-Exp3} and present it in Algorithm~\ref{alg:q_GDRO:col}.
\begin{algorithm}[t]
	\caption{\textsc{NPLA-Exp3} for non-oblivious PLA}
	\textbf{Input}: a scheduler $\A_s$
	\begin{algorithmic}[1]
		\STATE Initialize $\Lb_{0}=\mathbf{0}_m$
		\FOR{$t=1,2,\dots$}
		\STATE Receive $r_t$ from $\A_s$ and compute $\q_t$ according to \eqref{eqn:GDRO:qt_update:col}
		\IF{$r_t=1$}
		\STATE Select $c_t\in[m]$ according to $\q_t$ and set $\I_t=\emptyset$ 
		\ELSE 
		\STATE Select $c_t\in[m]$ according to $\q_t$ and generate $\I_t=\text{DepRound}(\frac{r_t-1}{m-1}\mathbf{1}_{m-1})$ from $[m]\backslash \{c_t\}$  
		\ENDIF
		\STATE Send $C_t=\{c_t\}\cup \I_t$, $c_t$ and $\q_t$ to $\A_s$
		\STATE Receive losses $\hats_{t,i}$ for all $i\in C_t$ from $\A_s$
		\STATE Construct $\tildesb_t$ in \eqref{eqn:GDRO:loss_estimator} and update $\Lb_{t}$ according to \eqref{eqn:GDRO:cumulative_loss_estimator:col}
		\ENDFOR
	\end{algorithmic}\label{alg:q_GDRO:col}
\end{algorithm}

We now state the theoretical guarantee for Algorithm~\ref{alg:q_GDRO:col}.
Although in rounds with $r_t\ge 2$ we employ the same loss estimator as~\citet{pmlr-v32-seldin14}, their analysis does not apply to our setting since it establishes only an expected regret guarantee against an \emph{oblivious} adversary, whereas our goal is to derive a high-probability bound in the more challenging \emph{non-oblivious} setting.
Moreover, \textsc{NPLA-Exp3} alternates between two loss estimators: 
on the one hand, we need to control the bias induced by implicit exploration;
on the other hand, it is also necessary to exploit the low variance offered by the multi-sample loss estimator.
We decompose the error according to whether $r_t=1$ or $r_t\ge 2$, control the two contributions via different analytical tools, and then carefully combine them to obtain an $r_t$-dependent regret bound.
The result is summarized in Lemma~\ref{lem:GDRO_q:col}, which provides, to the best of our knowledge, the \emph{first} high-probability regret bound for non-oblivious PLA.
\begin{lemma} \label{lem:GDRO_q:col} 
    Let $\eta_{q,t}$ and $\gamma_t$ be defined in \eqref{eqn:proof:q:stepsize:1} and~\eqref{eqn:proof:q:stepsize:sep:gammat} for \textsc{NPLA-EXP3}. 
    Under Assumption~\ref{ass:gen_loss_bounded}, for any fixed $t\in\N$ and any  sample-size process $\{r_j\}_{j\in[t]}$ generated according to Definition~\ref{def:sample_size_protocol}, with probability at least $1-\delta$,
	\begin{equation*}\label{eqn:lem:GDRO_q:High-pro:col}
    \Regqp=O\left(\sqrt{\sum_{j=1}^t \frac{m}{r_j}}\left(\sqrt{\log m}+\frac{\log \frac{1}{\delta}}{\sqrt{\log m}}\right)\right).
	\end{equation*}
\end{lemma}
\textbf{Remark:} The above high-probability regret bound for non-oblivious PLA matches the order of the expected regret bound in the oblivious setting~\citep{pmlr-v32-seldin14}.
For fixed $r_t\equiv r<m$, it achieves the optimal polynomial dependence on $t$ and $m/r$ up to logarithmic factors, since the lower bound for oblivious PLA \citep[Theorem~2]{pmlr-v32-seldin14} also applies to our setting. At the full-information endpoint $r=m$, the bound reduces to $O(\sqrt{t\log m})$, which matches the minimax optimal dependence for prediction with expert advice \citep[Table~1]{INF}.

\subsection{FTRL for the $w$-player}  \label{sec:w-player}
According to the decomposition in~\eqref{eqn:decom:1}, we need to control the standard regret of the $w$-player. To obtain an anytime guarantee, one may follow prior work and use SMD with time-varying step sizes. However, existing analyses only control a weighted average regret~\citep{zhang2026-GDRO-TPAMI}, while the standard regret may become linear under Assumption~\ref{ass:gen_domain_w}~\citep{ORABONA201850}. We therefore adopt FTRL with time-varying step sizes, which yields the desired standard regret guarantee.

To implement FTRL, we maintain the cumulative gradient vector $\Fb_t \in \R^d$, where $d$ is the dimension of $\W$, and initialize $\Fb_0=\mathbf{0}_d$. In each round $t$, we update $\w_t$ as
\begin{equation}\label{eqn:update:wt}
	\w_t= \argmin_{\w \in \W} \Big\{ \Big\langle\eta_{w,t}\Fb_{t-1},\w \Big\rangle + \nu_w\left(\w\right) \Big\},
\end{equation}
where $\eta_{w,t}>0$ is the step size.
If we set $\nu_w(\w)=\frac{1}{2}\|\w\|_2^2$, then \eqref{eqn:update:wt} has a closed-form solution $\w_t= \Pi_{\W} \left[ -\eta_{w,t}\Fb_{t-1} \right]$, where $\Pi_{\W}[\cdot]$ denotes the Euclidean projection onto $\W$.
To construct an unbiased stochastic gradient for the $w$-player, we can reuse the sample $\z_t^{(c_t)} \sim \P_{c_t}$ queried by the $q$-player. 
Specifically, the $w$-player sends\footnote{The ``send'' operation is notational and incurs no communication cost.} $\w_t$ to $\A_s$, which then returns a stochastic gradient
\begin{equation}  \label{eqn:GDRO:gw}
	\g_w(\w_t,\q_t)=\nabla\ell\left(\w_t;\z_t^{(c_t)}\right).
\end{equation}
Under Assumption~\ref{ass:gen_gradient_bounded}, $\g_w(\w_t,\q_t)$ is bounded, and since $c_t \sim \q_t$, it is an unbiased estimator of $\nabla_\w\phi\left(\w_t,\q_t\right)$: 
\begin{equation} \label{eqn:update:wt:unbiased_gradient}
    \left\|\g_w(\w_t,\q_t)\right\|_{w,*} \overset{\eqref{eqn:ass:gradient_bound}}{\leq} G, \
    \E_{t-1}\left[\g_w(\w_t,\q_t)\right]=\nabla_\w\phi\left(\w_t,\q_t\right),
\end{equation}
where $\E_{t-1}[\cdot]$ denotes the expectation conditioned on the randomness up to round $t-1$ and the value of $r_t$ in round $t$. Then, we update $\Fb_t$ as 
\begin{equation}\label{eqn:upate:Fb}
	\Fb_t=\Fb_{t-1}+\g_w(\w_t,\q_t).
\end{equation}
Algorithm~\ref{alg:w} summarizes FTRL for non-oblivious OCO.
\begin{algorithm}[t]
	\caption{FTRL for non-oblivious OCO}
	\textbf{Input}: a scheduler $\A_s$
	\begin{algorithmic}[1]
		\STATE Initialize $\Fb_0=\mathbf{0}_d$
		\FOR{$t=1,2,\dots$}
		\STATE Update $\w_t$ according to \eqref{eqn:update:wt} and send $\w_t$ to $\A_s$
		\STATE Receive $\g_w(\w_t,\q_t)$ in \eqref{eqn:GDRO:gw} from $\A_s$ and update $\Fb_t$ according to \eqref{eqn:upate:Fb}
		\ENDFOR
	\end{algorithmic}\label{alg:w}
\end{algorithm}

We then establish the regret bound for Algorithm~\ref{alg:w} in Lemma~\ref{lem:wt}. To address the challenge of non-obliviousness, we incorporate the ``ghost iterate'' technique introduced by \citet{nemirovski-2008-robust} into the FTRL analysis.
\begin{lemma} \label{lem:wt}
	Under Assumptions~\ref{ass:gen_domain_w}, \ref{ass:gen_gradient_bounded} and \ref{ass:gen_loss_convex}, and setting $\eta_{w,t}=\frac{\sqrt{2}D}{\sqrt{5}G}\frac{1}{\sqrt{t}}$ for Algorithm~\ref{alg:w}, we have for any fixed $t\in\N$, with probability at least $1-\delta$,
	\begin{equation} \label{eqn:lem:wt}
		\Regw \leq DG\sqrt{t}\left(2\sqrt{10}+8\sqrt{\ln \frac{1}{\delta}}\right).
	\end{equation}
\end{lemma}
\textbf{Remark:}
The high-probability regret bound \eqref{eqn:lem:wt} is of the same order as the non-anytime bound established by \citet[Theorem 5]{zhang2023-SA-GDRO}. Due to different definitions of regret, our result is not comparable to the anytime weighted average regret bound of~\citet{zhang2026-GDRO-TPAMI}.

\begin{algorithm}[t]
	\caption{Meta-scheduler for GDRO with FSQ}
    \textbf{Instantiate} the OCO subroutine $\A_w$ and the PLA subroutine $\A_q$, and run them in parallel.
    
	\begin{algorithmic}[1]
		\FOR{$t=1,2,\dots$}
		\STATE Receive $r_t$ from the environment and send $r_t$ to $\A_q$
		\STATE Receive $C_t$, $c_t$, and $\q_t$ from $\A_q$
		\STATE Receive $\w_t$ from $\A_w$
		\STATE Draw samples $\{\z_t^{(i)}\}_{i\in C_t}$
        \STATE Construct and send $\hats_{t,i}$ for all $i\in C_t$  to $\A_q$
        \STATE Construct and send $\nabla\ell(\w_t;\z_t^{(c_t)})$  to $\A_w$
		\STATE Compute the averages $\wb_t$ and $\qb_t$ in \eqref{eqn:average_output}
		\ENDFOR
	\end{algorithmic}\label{alg:all_GDRO_two_layer}
\end{algorithm}

\subsection{The meta-scheduler and \textsc{Flex-GDRO}}\label{sec:player_q_strategy_col}
We present the meta-scheduler $\A_s$ in Algorithm~\ref{alg:all_GDRO_two_layer}, which obtains $r_t$ from the environment and draws samples for the updates of both players. 
By instantiating $\A_q$ and $\A_w$ in $\A_s$ as \textsc{NPLA-Exp3} (Algorithm~\ref{alg:q_GDRO:col}) and \textsc{FTRL} (Algorithm~\ref{alg:w}), respectively, we obtain \textsc{Flex-GDRO}, an anytime algorithm that can process an arbitrary number of samples per round. 
We discuss its computational cost in Appendix~\ref{app:practical_overhead}.
Combining Lemmas~\ref{lem:GDRO_q:col} and~\ref{lem:wt} with the decomposition in~\eqref{eqn:decom:2}, we establish the following optimization error bound.
\begin{theorem} \label{thm:GDRO:col} 
    For \textsc{Flex-GDRO} with the step sizes in Lemmas~\ref{lem:GDRO_q:col} and~\ref{lem:wt}, under Assumptions~\ref{ass:gen_domain_w}--\ref{ass:gen_loss_convex}, for any fixed $t\in\N$ and any sample-size process $\{r_j\}_{j\in[t]}$ generated according to Definition~\ref{def:sample_size_protocol}, with probability at least $1-\delta$,
	\begin{equation}\label{eqn:cor:GDRO_all:high}
        \epsilon_{\phi}(\wb_t, \qb_t)=O\left(\frac{1}{t}\sqrt{\sum_{j=1}^t \frac{m}{r_j}}\left(\sqrt{\log m}+\frac{\log \frac{1}{\delta}}{\sqrt{\log m}}\right)\right).
	\end{equation}
\end{theorem}

\noindent\textbf{Remark:}
The high-probability bound in~\eqref{eqn:cor:GDRO_all:high} does not require knowing the total number of rounds in advance.
In practice, $r_t$ may be revealed online according to available resources. Our result accommodates this setting and shows that larger sample sizes lead to smaller optimization error.

\noindent\textbf{Remark:}
When the sample size is fixed, namely $r_j \equiv r$ for all $j\in[t]$, the upper bound~\eqref{eqn:cor:GDRO_all:high} simplifies to $O(\sqrt{m \log(m) / (r t)})$, implying a sample complexity of $O(m\log (m)/\epsilon^2)$ for all $r\in[m]$.
This matches the sample complexity of non-anytime GDRO algorithms in the extreme cases $r=1$~\citep{zhang2023-SA-GDRO} and $r=m$~\citep{nemirovski-2008-robust}, and nearly attains the lower bound $\Omega(m/\epsilon^2)$~\citep{DRO:Online:Game}.\footnote{This lower bound applies here because it is stated in terms of the total number of stochastic-oracle calls.}
Moreover, our rate improves upon the SMD-based anytime guarantees~\citep{zhang2026-GDRO-TPAMI} by a factor of $O(\log t)$, which stems from using FTRL with the averaged outputs~\eqref{eqn:average_output}, thus avoiding the accumulation of $\{1/j\}_{j=1}^t$.
Finally, the sample size is extended from $r\in\{1,m\}$ to all $r\in[m]$.

We further strengthen Theorem~\ref{thm:GDRO:col} by deriving a time-uniform bound that holds for all $t\in\N$.
\begin{theorem} \label{thm:GDRO:col_time_uniform} 
    Let \textsc{Flex-GDRO} be run with the step sizes specified in Lemmas~\ref{lem:GDRO_q:col} and~\ref{lem:wt}. Let $\{r_t\}_{t\ge 1}$ be any  sample-size process generated according to Definition~\ref{def:sample_size_protocol}. Under Assumptions~\ref{ass:gen_domain_w}--\ref{ass:gen_loss_convex}, with probability at least $1-\delta$, the following bound holds simultaneously for all $t\in\N$:
	\begin{equation*}\label{eqn:cor:GDRO_all:high_time_uniform}
		\epsilon_{\phi}(\wb_t, \qb_t)=O\left(\frac{1}{t}\sqrt{\sum_{j=1}^t \frac{m}{r_j}}\left(\sqrt{\log m}+\frac{\log\frac{\log t}{\delta}}{\sqrt{\log m}}\right)\right).
	\end{equation*}
\end{theorem}
\textbf{Remark:} 
The time-uniform guarantee incurs a minor additional cost compared to the fixed-time bound in~\eqref{eqn:cor:GDRO_all:high}, increasing the factor from $O(\sqrt{\log m})$ to $O(\max\{\sqrt{\log m},\log\log (t)/\sqrt{\log m}\})$. 
Compared with taking a union bound over all $t$ with varying confidence levels, which incurs an $O(\sqrt{\log t})$ cost, our method leverages time-uniform concentration inequalities~\citep{lee2025lasso,lee2025minimax} to reduce the cost to $O(\sqrt{\log\log t})$.

\begin{figure*}[t]
	\centering
	\begin{minipage}[t]{0.48\textwidth}
		\centering
		\subfigure[The synthetic dataset]{
			\label{fig:GDRO_varying_time_synthetic}
			\includegraphics[width=0.465\textwidth]{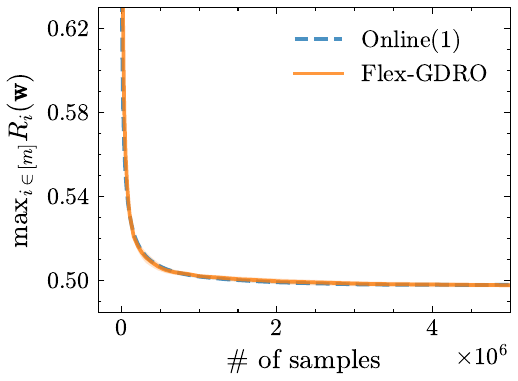}
		}
		\subfigure[The diabetes dataset]{
			\label{fig:GDRO_varying_time_diabetes}
			\includegraphics[width=0.465\textwidth]{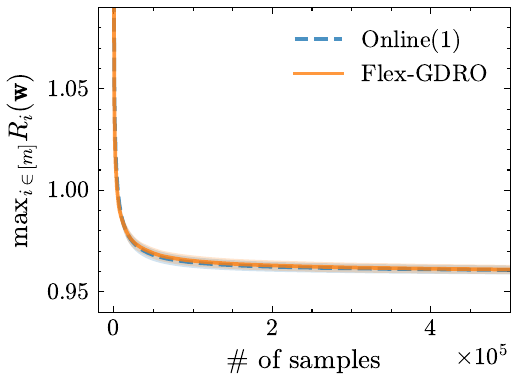}
		}
		\caption{Dynamic-resource setting: the maximum risk versus the number of samples.}
		\label{fig:GDRO_varying_sample}
	\end{minipage}
	\hfill
    \begin{minipage}[t]{0.48\textwidth}
		\centering
		\subfigure[The synthetic dataset]{
			\label{fig:GDRO_varying_risk_synthetic}
			\includegraphics[width=0.465\textwidth]{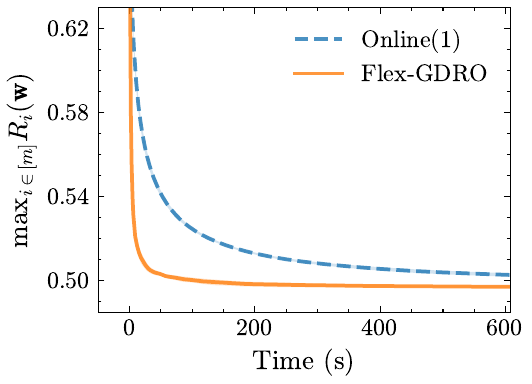}
		}
		\subfigure[The diabetes dataset]{
			\label{fig:GDRO_varying_risk_diabetes}
			\includegraphics[width=0.465\textwidth]{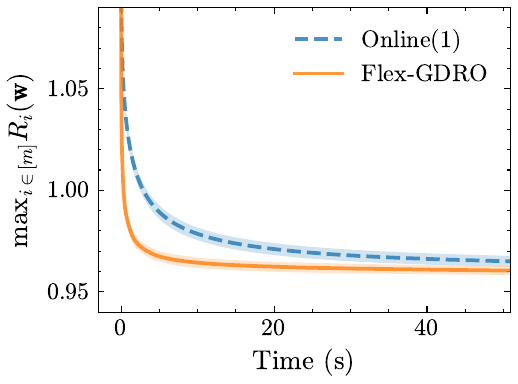}
		}
		\caption{Dynamic-resource setting: the maximum risk versus the running time.}
		\label{fig:GDRO_varying_time}
	\end{minipage}
\end{figure*}

\begin{table*}[t]
\renewcommand{\arraystretch}{1.14}
\caption{Comparison of \textsc{Online}($1$) and \textsc{Flex-GDRO} on the synthetic and diabetes datasets. The experiments use five independent runs. \textit{Time} is measured in seconds. \textit{Speedup} is the runtime ratio of \textsc{Online}($1$) to \textsc{Flex-GDRO} for achieving the same risk. Gray shading highlights \textsc{Flex-GDRO}.}
\label{tab:online-vs-FSQ}
\small
\setlength{\tabcolsep}{3pt}
\begin{center}
\input{tables/synthetic_diabetes.tex}
\end{center}
\end{table*}
\begin{figure*}[t]
	\centering
	\begin{minipage}[t]{0.48\textwidth}
	\centering
    \subfigure[The synthetic dataset]{
        \label{fig:FSQ-U_synthetic_round}
        \includegraphics[width=0.467\textwidth]{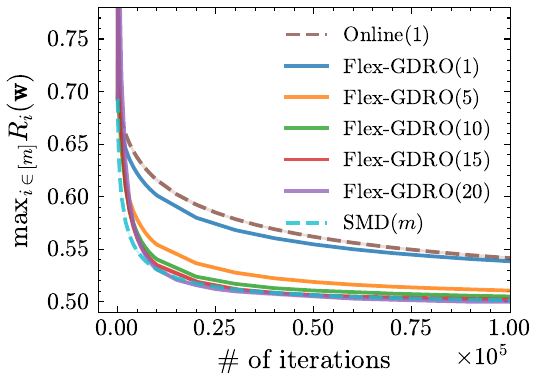}
    }
    \subfigure[The diabetes dataset]{
        \label{fig:FSQ-U_diabetes_round}
        \includegraphics[width=0.467\textwidth]{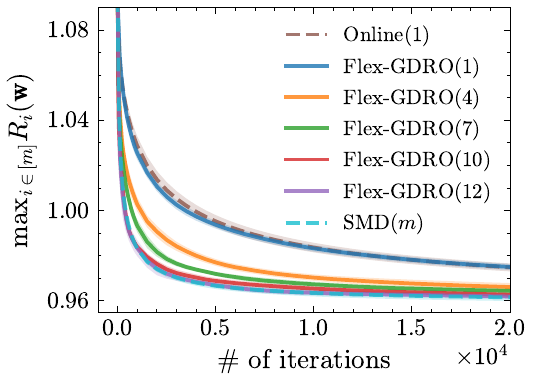}
    }
	\vspace{-0.3em}
	\caption{Fixed-resource setting: the maximum risk versus the number of iterations.}
	\label{fig:FSQ_round}
	\end{minipage}
	\hfill
	\begin{minipage}[t]{0.49\textwidth}
		\centering
		\subfigure[The synthetic dataset]{
			\label{fig:FSQ-U_synthetic_sample}
			\includegraphics[height=0.1\textheight]{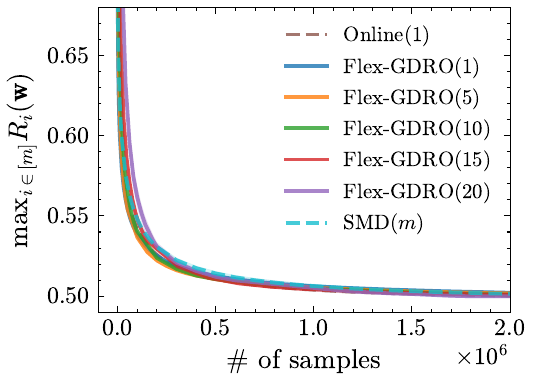}
		}
		\subfigure[The diabetes dataset]{
			\label{fig:FSQ-U_diabetes_sample}
			\includegraphics[height=0.1\textheight]{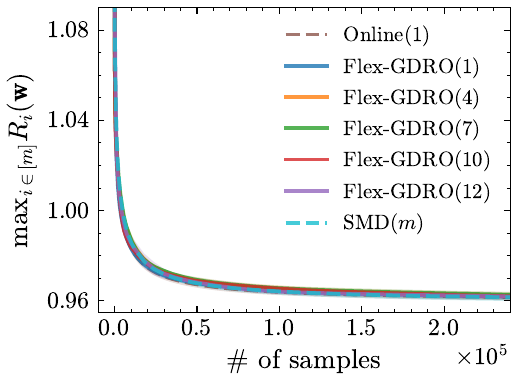}
		}
		\caption{Fixed-resource setting: the maximum risk versus the number of samples.}
		\label{fig:FSQ_sample}
	\end{minipage}
\end{figure*}

\section{Experiments}  \label{sec:experiments:main}
In this section, we evaluate the sample and time efficiency of \textsc{Flex-GDRO} in dynamic-resource and fixed-resource settings and examine the variance of its loss estimator.

\subsection{Experimental Setup}
We consider four datasets: a synthetic binary dataset ($m=20$), a multi-class diabetes dataset ($m=12$)~\citep{diabetes_dataset}, CivilComments-WILDS ($m=16$)~\citep{2021_WILDS}, and FEMNIST ($m=100$)~\citep{2017_IJCNN_EMNIST,2019_LEAF}. We evaluate \textsc{Flex-GDRO} on all four datasets under the dynamic-resource setting with $r_t\in[m-1]$, and additionally consider the fixed-resource setting with $r\in[m]$ on the synthetic and diabetes datasets. We also conduct synthetic experiments with larger numbers of groups to examine the variance of the loss estimator of \textsc{Flex-GDRO}.

All tasks train a linear classifier with cross-entropy loss. We compare \textsc{Flex-GDRO} (Algorithm~\ref{alg:all_GDRO_two_layer} instantiated with Algorithms~\ref{alg:q_GDRO:col} and~\ref{alg:w}) with \textsc{SMD}($m$) and \textsc{Online}($1$), which are anytime variants of Algorithms~1 and~2 in~\citet{zhang2026-GDRO-TPAMI} and use $m$ samples and $1$ sample per round, respectively.
For the synthetic and diabetes datasets, we report mean risk curves, with standard deviations shown as shaded regions. Further experimental details are provided in Appendix~\ref{app:experiments}.

\subsection{Dynamic-resource Scenario}
We compare \textsc{Flex-GDRO} with \textsc{Online}($1$) on all four datasets under varying per-round sample budgets, where \textsc{SMD}($m$) is not applicable without a buffering scheme.

For the synthetic and diabetes datasets, we uniformly sample $r_t\in[1, m-1]$ in each round.
Figs.~\ref{fig:GDRO_varying_sample} and~\ref{fig:GDRO_varying_time} present the maximum risk versus the number of samples and the running time, respectively.
As shown in Fig.~\ref{fig:GDRO_varying_sample}, the risk curves of \textsc{Flex-GDRO} and \textsc{Online}($1$) are closely aligned, suggesting similar sample complexities.
Fig.~\ref{fig:GDRO_varying_time} demonstrates that the risk of \textsc{Flex-GDRO} decreases more rapidly than that of \textsc{Online}($1$). This is because \textsc{Flex-GDRO} processes multiple samples per update, whereas \textsc{Online}($1$) is restricted to $1$ sample. As a result, given a comparable sample complexity, \textsc{Flex-GDRO} reaches the target risk in fewer updates and therefore in less wall-clock time.
Table~\ref{tab:online-vs-FSQ} further reports the runtime required to reach the same risk, showing that \textsc{Online}($1$) requires approximately 5 to 13 times longer than \textsc{Flex-GDRO}. These results highlight the advantage of processing multiple samples simultaneously rather than handling them one by one.

On CivilComments-WILDS, we freeze a DistilBERT~\citep{2019_DistilBERT} encoder, and train only a linear classification head. We consider $m=16$ groups and uniformly sample $r_t\in[1,m-1]$ in each round. 
Table~\ref{tab:civilcomments} reports results at five target risks. Across these thresholds, \textsc{Online}($1$) takes 8.40 to 11.20 times longer than \textsc{Flex-GDRO}. Moreover, \textsc{Flex-GDRO} uses fewer samples at risks $0.65$ and $0.64$. Implementation details are provided in Appendix~\ref{app:civilcomments}.

\begin{table}[ht]
	\renewcommand{\arraystretch}{1.14}
    \caption{Comparison of \textsc{Online}($1$) and \textsc{Flex-GDRO} on CivilComments-WILDS. \textit{Time} is measured in seconds. \textit{Speedup} is the runtime ratio of \textsc{Online}($1$) to \textsc{Flex-GDRO} for achieving the same risk. Gray shading highlights \textsc{Flex-GDRO}.}
    \label{tab:civilcomments}
    \small
    \setlength{\tabcolsep}{6pt}
    \centering
    \input{tables/civilcomments.tex}
\end{table}

We further evaluate \textsc{Flex-GDRO} on FEMNIST with $100$ writer groups under dynamic resources. At a maximum group risk of $0.94$, it uses nearly the same number of samples as \textsc{Online}($1$) while requiring less optimization time. Detailed setups and results are provided in Appendix~\ref{app:femnist}.

\subsection{Fixed-resource Scenario}
We consider a scenario with a fixed number of samples $r\in\{1,5,10,15,20\}$ for the synthetic dataset and $r\in\{1,4,7,10,12\}$ for the diabetes dataset. For brevity, we denote \textsc{Flex-GDRO} with $r$ samples as \textsc{Flex-GDRO}($r$). 
Figs.~\ref{fig:FSQ_round} and~\ref{fig:FSQ_sample} report the maximum risk versus the number of iterations and the  number of queried samples, respectively.
As shown in Fig.~\ref{fig:FSQ_round}, larger $r$ leads to faster risk reduction, which is consistent with the optimization error bound of $O(\sqrt{m\log (m)/(rt)})$ for any fixed $r\in[m]$.
Fig.~\ref{fig:FSQ_sample} shows that the risk curves of \textsc{Flex-GDRO}($r$) nearly overlap across different values of $r$, in line with the theoretical sample complexity of $O(m\log (m)/\epsilon^2)$. 
Moreover, in Fig.~\ref{fig:FSQ_sample}, the risk curves of \textsc{SMD}($m$) and \textsc{Online}($1$) closely match those of \textsc{Flex-GDRO}($r$) across different values of $r$, indicating similar sample complexities.

\subsection{Estimator Variance}
For synthetic experiments with $m\in\{50,100,200\}$ and fixed ratios $m/r\in\{5,10,25\}$, the $\q_t$-weighted variance of the loss estimator remains similar as $m$ increases at a fixed ratio, and increases with $m/r$. This supports the conclusion that the regret-relevant variance is governed primarily by $m/r_t$. Appendix~\ref{app:variance} reports the estimation protocol and the results.

\section{Conclusion and future work}  \label{sec:conclusion}
This paper studies GDRO in a novel FSQ setting, where the per-round sample size may vary over time. 
We formulate the problem as a two-player game, with one player solving non-oblivious OCO via FTRL and the other solving non-oblivious PLA via our proposed \textsc{NPLA-Exp3}. 
By integrating the two components, we develop \textsc{Flex-GDRO}, an anytime GDRO algorithm that supports FSQ, and derive optimization error bounds that depend on the per-round sample sizes.
Future work could extend FSQ to other optimization formulations, such as MRO~\citep{MRO-tongzhang,MERO}, and explore its applications to federated learning~\citep{2019_Agnostic_FL,2020_DRFA,2026_AFLCom}, recommendation systems~\citep{Wang_Jian_Yang_Lu_Shen_Wang_Zeng_Zhang_2025}, and large language models~\citep{GDRO_LLM_app_pretraing,qiu2024ICML,GDRO_LLM_app1,2026_GDRO_LLM_app_Tencent}.

\begin{ack}
This work was partially supported by National Science and Technology Major Project (2022ZD0114801), NSFC (62361146852), the ``111 Center'' (No. B26023), and the Collaborative Innovation Center of Novel Software Technology and Industrialization.

The authors declare no competing interests.
\end{ack}

\bibliographystyle{plainnat}
\bibliography{ref}

\appendix
\section{Representative Applications of FSQ} \label{sec:applications}
In this section, we present three representative applications of the FSQ setting. In all three applications, the per-round sample size $r_t$ is externally determined by potentially time-varying resource constraints, while the learner can still choose which groups to query.

\paragraph{Federated learning in mobile edge computing (MEC).}
In MEC-based federated learning, client participation is coordinated by the  server through centralized resource allocation~\citep{2019_FedCS,2024_CS_Survey}. 
To improve robustness and fairness across clients, one may adopt agnostic federated learning~\citep{2019_Agnostic_FL}, whose objective coincides with GDRO when each client is treated as a group. 
Since wireless resources are limited and communication conditions may fluctuate over time, for example due to bandwidth variations~\citep{ZHANG2020242}, only a finite and potentially time-varying number of clients can participate in each round.
In this case, we view the number of clients that the server can query in round $t$ as $r_t$.

\paragraph{Multilingual content moderation.}
In multilingual content moderation, the massive volume of online content makes an efficient moderation system essential~\citep{2023_Multilingual_Reddit,2023_FairM_KDD}. A practical goal is to train a model that predicts whether content complies with platform policies. 
In this setting, human judgment is indispensable for providing high-quality labels for model improvement. Since data availability differs substantially across languages, especially between English and other languages, it is necessary to balance overall accuracy and fairness~\citep{2023_Multilingual_Reddit}. Under this objective, a GDRO formulation over language groups is well motivated. Meanwhile, human review capacity is limited and time-varying due to changing workloads, reviewer/partner capacities, and staffing requirements~\citep{2023_FairM_KDD}. We therefore model the number of labels that can be queried in round $t$ as $r_t$, and let the platform allocate these label queries across language groups. Given the large scale of the data, it is reasonable to assume that a sample is available from each group in every round.

\paragraph{Medical image analysis.}
In medical image analysis, the goal is to train a deep model from a pool of unlabeled medical images. When the model is trained using empirical risk minimization, performance may vary substantially across patient subgroups, raising concerns about subgroup disparity and fairness in downstream clinical use~\citep{2022_Medical_GDRO,2024_COVID_19}. It is therefore necessary to account for both overall predictive performance and subgroup robustness. Under this objective, applying GDRO over patient subgroups, such as those defined by race, sex, and age, is a natural choice~\citep{2022_Medical_GDRO}. Since expert annotation is expensive and time-consuming~\citep{2024_Survey_Medical}, only a limited number of annotations can be obtained in each round. We therefore model the annotation capacity in round $t$ as $r_t$, and allow it to vary over time to reflect changing operating conditions. Given that the unlabeled pool is typically much larger than the available expert capacity, it is reasonable to assume that each group has an available instance in every round.

\section{Related Work}  \label{app:related_work}
In this section, we briefly review GDRO and PLA.

\subsection{Group Distributionally Robust Optimization}
GDRO is motivated by applications affected by distribution shift, including federated learning~\citep{2019_Agnostic_FL}, language modeling~\citep{GDRO-oren2019}, and neural network training~\citep{Gouop_DRO_ICLR_2020}.
A key algorithmic objective is to minimize the \emph{sample complexity}, i.e., the number of samples required to attain an $\epsilon$-optimal solution. 
\citet{DRO:Online:Game} establish a lower bound of $\Omega(m/\epsilon^2)$, and state-of-the-art methods achieve a near-optimal bound of $O(m\log(m)/\epsilon^2)$.
These methods fall into two sampling paradigms:
(i) $m$-sample methods that query one sample from each of the $m$ distributions per round, e.g., stochastic mirror descent (SMD)~\citep{nemirovski-2008-robust};
(ii) 1-sample methods that draw a single sample per round, including gradient-clipped SMD~\citep{NEURIPS2022} and online-learning-based two-player game formulations~\citep{Online:Multiple:Distribution,DRO:Online:Game,zhang2023-SA-GDRO}.
Beyond sample efficiency, another practical consideration is whether an algorithm is \emph{anytime}~\citep{pmlr-v97-cutkosky19a,pmlr-v32-luo14}. In contrast to the above non-anytime methods, which require the number of iterations to be specified in advance, \citet{zhang2026-GDRO-TPAMI} propose anytime GDRO algorithms with optimization error of $O(\log(t)\sqrt{m\log(m)/(rt)})$ for $r\in\{1,m\}$.
Overall, existing GDRO methods are restricted to the two extremes of processing either $1$ or $m$ samples per round, leaving the intermediate and variable sample sizes under-explored.

Several works also explore GDRO under additional assumptions or different formulations. 
\citet{nguyen2025beyond} exploit sparsity to reduce the dependence on $m$ in the sample complexity.
\citet{MRO-tongzhang} introduce minimax regret optimization (MRO) to prevent the worst-case objective from being dominated by a single distribution, which can be viewed as a special case of calibrated DRO~\citep{pmlr-v151-slowik22a}.  
Subsequent studies develop efficient algorithms for MRO in finite-distribution settings~\citep{MERO,ICML:2024:Yu} and algorithms for GDRO in decentralized settings~\citep{2026_DecGDRO}.

\subsection{Prediction with Limited Advice}
PLA~\citep{pmlr-v32-seldin14} can be regarded as an intermediate case between multi-armed bandit~\citep{Robbins-1952,Auer02}, where the player selects $1$ expert per round and observes its advice, and prediction with expert advice~\citep{Hedge_1997}, where the player observes the advice of all $m$ experts. 
In PLA, the player can query the advice of $r_t\in[m]$ experts in round $t$, and the loss is evaluated on a subset of the selected experts~\citep{pmlr-v32-seldin14,pmlr-v201-saad23a}.
\citet{pmlr-v32-seldin14} study oblivious PLA with varying sample sizes and establish an expected regret bound of $O\left(\sqrt{\sum_{j=1}^t \frac{m}{r_j}\log m}\right)$.
Subsequent work investigates PLA across multiple settings, including access to distributions over experts for selection~\citep{pmlr-v35-kale14a}, non-uniform expert costs~\citep{Amin_Kale_Tesauro_Turaga_2015}, fixed per-round sample sizes~\citep{Twolayers-ACM-2018}, stochastic loss sequences, or adversarial ones with a small loss range~\citep{NEURIPS2018_253f7b5d}, excess generalization error in stochastic settings~\citep{NEURIPS2021_c688defd}, and convex-combination predictions over experts~\citep{pmlr-v201-saad23a}.

Despite extensive studies, existing PLA algorithms are inapplicable to our setting, which requires minimizing regret against a non-oblivious adversary with varying sample sizes, because prior work assumes stochastic adversaries~\citep{NEURIPS2021_c688defd}, oblivious adversaries~\citep{pmlr-v32-seldin14,pmlr-v201-saad23a},  fixed sample sizes~\citep{NEURIPS2018_253f7b5d,Twolayers-ACM-2018} or focuses on pseudo-regret objectives~\citep{pmlr-v35-kale14a,Amin_Kale_Tesauro_Turaga_2015}.

Beyond PLA, another line of work allowing multiple observations per round studies graph-structured partial feedback, where selecting one action may reveal the losses of neighboring actions through a
feedback graph~\citep{NIPS2011_e1e32e23,pmlr-v40-Alon15,lu2021stochastic_1,lu2021stochastic_2,pmlr-v162-kong22b,pmlr-v201-luo23a,pmlr-v235-zhang24ce}.

\section{Experimental Details} \label{app:experiments}
All the experiments are implemented in Python and conducted on a server running Ubuntu 20.04, equipped with dual Intel(R) Xeon(R) Platinum 8358P CPUs (2.60 GHz).

During training, synthetic samples are generated on the fly, while samples from diabetes, CivilComments-WILDS, and FEMNIST are drawn uniformly with replacement within each queried group. During evaluation, we estimate each group's expected loss by an empirical average and report the maximum over groups.
Since \textsc{NPLA-Exp3} requires bounded loss feedback, we clip and normalize the cross-entropy loss used in the \(q\)-player update in all experiments. Let \(\ell_{\mathrm{ce}}(w;z)\) denote the cross-entropy loss. We define
\[
\ell_C(w;z)=\frac{\min\{\ell_{\mathrm{ce}}(w;z),C\}}{C},
\]
where the clipping threshold is set to \(C=5\). The \(q\)-player then forms the surrogate loss as $\hat{s}_{t,i}=1-\ell_C(w_t;z_t^{(i)})$, so its feedback lies in \([0,1]\). For the \(w\)-player update and evaluation, we use the original, unclipped cross-entropy loss.

\paragraph{Data and model licenses.}
Diabetes 130-US Hospitals~\citep{diabetes_dataset}\footnote{\url{https://archive.ics.uci.edu/dataset/296/diabetes+130-us+hospitals+for+years+1999-2008}}, CivilComments-WILDS~\citep{2021_WILDS}\footnote{\url{https://wilds.stanford.edu/datasets/\#civilcomments}}, and the \texttt{distilbert-base-uncased} checkpoint~\citep{2019_DistilBERT}\footnote{\url{https://huggingface.co/distilbert/distilbert-base-uncased}} are distributed under CC BY 4.0, CC0 1.0, and Apache 2.0, respectively. Federated EMNIST~\citep{2017_IJCNN_EMNIST,2019_LEAF}\footnote{\url{https://www.tensorflow.org/federated/api_docs/python/tff/simulation/datasets/emnist}} is provided through TensorFlow Federated and is derived from the LEAF preprocessing of EMNIST.
The TensorFlow Federated and LEAF codebases are released under Apache 2.0 and BSD 2-Clause, respectively.

\subsection{Synthetic and Diabetes Datasets}
We construct the synthetic dataset following prior work \citep{NIPS2016_4588e674, DRO:Online:Game, zhang2023-SA-GDRO, MERO,zhang2026-GDRO-TPAMI}. Specifically, we consider $m=20$ distributions, each associated with a ground-truth classifier $\w_i^*\in\R^{500}$. The set $\{\w_i^*\}_{i \in [m]}$ is constructed following the approach outlined for the synthetic dataset by \citet{zhang2026-GDRO-TPAMI}. For each distribution, we randomly generate samples $(\x,y)$, where $\x$ is drawn from the standard normal distribution $\mathcal{N}(0, I)$ and $y$ is generated as $y = \text{sign}(\x^{\top} \w_i^*)$ with probability $0.9$, or $y = -\text{sign}(\x^{\top} \w_i^*)$ with probability $0.1$.

For diabetes, we employ the dataset of \citet{diabetes_dataset} to perform a multi-class classification task. The goal is to predict a patient’s hospital readmission status, categorized as readmitted within 30 days, readmitted after 30 days, or not readmitted, based on treatment-related features. The original dataset contains 101,766 instances and 47 attributes. After preprocessing by removing three features with predominantly missing values, three highly sparse features unsuitable for one-hot encoding, and two irrelevant features, we obtain a refined dataset of 99,492 instances with no missing values and a 102-dimensional expanded feature representation. We partition the dataset into $m=12$ groups based on the combination of three sensitive attributes: race \{Caucasian, African American, Other\}, age \{$\leq60$, $>60$\}, and gender \{Male, Female\}. Each distribution $\P_i$ corresponds to the empirical distribution of the samples in the $i$-th group.

\subsection{CivilComments-WILDS} \label{app:civilcomments}
We use the official training, validation, and test splits of CivilComments-WILDS. The task is binary toxicity classification, with groups formed by crossing eight identity attributes with the binary toxicity label. We use the frozen \texttt{distilbert-base-uncased} encoder and precompute its representations once. The reported wall-clock time measures only the subsequent optimization of the linear classification head.
The reported run uses $30{,}000$ rounds. Both methods use the same initialization and $r_t$ sequence, with \textsc{Online}($1$) processing samples sequentially. Hyperparameters for each method are selected from the same fixed candidate set using validation performance.

\subsection{FEMNIST with 100 Writer Groups} \label{app:femnist}
For the FEMNIST experiments, we use the TensorFlow Federated (TFF) partition of EMNIST for $62$-class handwritten-character classification~\citep{2017_IJCNN_EMNIST,2019_LEAF}. The \texttt{train} and \texttt{test} splits contain the same $3{,}400$ writers, with $671{,}585$ training samples and $77{,}483$ test samples. We rank writers by their number of training samples and select the top $100$ as the target groups. 
The non-target pool consists of all $3{,}300$ remaining writers, which we split into $2{,}970$ writers for CNN training and $330$ writers for CNN validation, corresponding to a $90\%/10\%$ split at the writer level. We pretrain a two-layer CNN using ERM on the $2{,}970$ writers. After removing the final classification layer, we freeze the encoder and train only a linear softmax head on the target groups with cross-entropy loss, retaining a convex optimization objective for the head.

We run \textsc{Flex-GDRO} for $10{,}000$ rounds, with $r_t=\lceil100\alpha_t\rceil$ and $\alpha_t$ drawn uniformly from $\{0.02,0.05,0.10,0.20\}$. We report maximum test risk over the target writers. Table~\ref{tab:femnist} gives the mean and sample standard deviation. 
At risk $0.94$, the methods require nearly the same number of samples, while \textsc{Flex-GDRO} takes less optimization time. At risk $0.93$, it uses fewer samples as well. The optimization-time ratios range from $3.32$ to $4.31$ across the reported thresholds. 
These results show that \textsc{Flex-GDRO} maintains competitive sample efficiency and a clear optimization-time advantage with $100$ groups.

\begin{table}[t]
	\renewcommand{\arraystretch}{1.14}
    \caption{Comparison of \textsc{Online}($1$) and \textsc{Flex-GDRO} on FEMNIST with $100$ writer groups under dynamic resources. Entries are mean $\pm$ sample standard deviation over five seeds at the first saved checkpoint reaching each threshold. \textit{Time} is measured in seconds for head optimization. \textit{Speedup} is the ratio of mean optimization times of \textsc{Online}($1$) to \textsc{Flex-GDRO} for achieving the same risk. Gray shading highlights \textsc{Flex-GDRO}.}
    \label{tab:femnist}
    \small
    \setlength{\tabcolsep}{8pt}
    \centering
    \input{tables/femnist.tex}
\end{table}

\subsection{Empirical Estimator Variance} \label{app:variance}
The adaptive loss estimator $\tildesb_t$ serves as the stochastic gradient of the $q$-player. Its entropy-FTRL regret is controlled by the $\q_t$-weighted second moment $M_{q,t}:=\sum_{i=1}^m q_{t,i}\mathbb{E}_{t-1}[\tilde{s}_{t,i}^2].$
For $r_t\geq2$, the estimator analysis in~\eqref{eqn:GDRO:loss_estimator} gives $M_{q,t}\leq \frac{m}{r_t}$. Hence, the corresponding $\q_t$-weighted variance $V_{q,t}:=\sum_{i=1}^m q_{t,i}\operatorname{Var}_{t-1}(\tilde{s}_{t,i})$ satisfies $V_{q,t}\leq M_{q,t}\leq m/r_t$. Thus, the variance quantity relevant to the regret analysis depends on the observation ratio $m/r_t$, rather than on $m$ alone.

To examine this dependence empirically, we conduct fixed-resource experiments on the synthetic dataset with $m\in\{50,100,200\}$ and $m/r\in\{5,10,25\}$. At checkpoints $t\in\{1{,}000,5{,}000,10{,}000,20{,}000\}$, we fix $(\w_t,\q_t)$ and generate $500$ independent realizations of $\tildesb_t$. We estimate the weighted variance by $\widehat V_{q,t}:=\sum_{i=1}^m q_{t,i}\widehat{\operatorname{Var}}(\tilde{s}_{t,i}),$ where $\widehat{\operatorname{Var}}(\tilde{s}_{t,i})$ denotes the empirical variance. Table~\ref{tab:variance-weighted} reports the mean and standard deviation over five independent seeds.

The results are consistent with the $O(m/r)$ dependence in analysis. For example, with $r=20$, increasing $m$ from $100$ to $200$ doubles $m/r$ from $5$ to $10$ and increases the weighted variance at $t=20{,}000$ from $2.703$ to $6.044$. By contrast, when $r$ grows proportionally with $m$ so that $m/r=5$, the  values are $2.688$, $2.703$, and $2.730$ for $m=50$, $100$, and $200$, respectively. These results indicate that the regret-relevant estimator variance is primarily determined by $m/r$.

We also consider the unweighted average per-coordinate variance $V_t:=\frac{1}{m}\sum_{i=1}^m\operatorname{Var}(\tilde{s}_{t,i}),$ with empirical estimate $\widehat V_t:=\frac{1}{m}\sum_{i=1}^m\widehat{\operatorname{Var}}(\tilde{s}_{t,i}).$ This quantity directly measures the average coordinate-wise fluctuation as $m$ increases. Table~\ref{tab:variance-unweighted} shows the same qualitative dependence on $m/r$.

Table~\ref{tab:variance-w} further reports the mean per-coordinate variance of the $w$-player stochastic gradient. Across all considered values of $m$ and $r$, the means at $t=20{,}000$ range from $0.172$ to $0.183$. This is consistent with our design, since the $w$-player uses a single gradient sampled according to $\q_t$, whose norm bound and regret guarantee are independent of the per-round sample size.

\begin{table}[t]
    \centering
    \caption{Empirical $\q_t$-weighted variance $\widehat V_{q,t}$ of the loss estimator on synthetic data. Entries are mean $\pm$ standard deviation over five seeds.}
    \label{tab:variance-weighted}
    \small
    \setlength{\tabcolsep}{4pt}
    \input{tables/variance_weighted.tex}
\end{table}

\begin{table}[t]
    \centering
    \caption{Unweighted mean per-coordinate variance $\widehat V_t$ of the loss estimator on synthetic data. Entries are mean $\pm$ standard deviation over five seeds.}
    \label{tab:variance-unweighted}
    \small
    \setlength{\tabcolsep}{4pt}
    \input{tables/variance_unweighted.tex}
\end{table}

\begin{table}[t]
    \centering
    \caption{Mean per-coordinate variance of the $w$-player stochastic gradient on synthetic data. Entries are mean $\pm$ standard deviation over five seeds.}
    \label{tab:variance-w}
    \small
    \setlength{\tabcolsep}{4pt}
    \input{tables/variance_w_gradient.tex}
\end{table}

\section{Limitations} \label{app:limitation}
Our FSQ framework focuses on settings where the available sample size may vary due to external resource constraints, while the learner is still allowed to decide which groups to query. This modeling captures applications such as centrally coordinated client selection, human review allocation, and expert annotation scheduling, as discussed in Appendix~\ref{sec:applications}. However, it does not cover settings in which workers or clients autonomously decide whether to join the learning process. Extending FSQ to such self-selected participation settings is an interesting direction for future work.

\section{Computational cost of \textsc{Flex-GDRO}} \label{app:practical_overhead}
The per-round computational cost of \textsc{Flex-GDRO} consists of three components: the $q$-player update, the $w$-player update, and group selection with sample acquisition. The first two components have the same order of complexity as existing GDRO methods, while the additional group-selection step requires $O(m)$ time per round. We detail these operations below.
\begin{compactenum}
    \item \textbf{$q$-player update.} The exponential-weights update~\eqref{eqn:GDRO:qt_update:col} requires $O(m)$ elementary operations. Accumulating the observed losses requires another $O(r_t)$ operations, so the overall cost is $O(m)$ since $r_t\leq m$. Entropy-based updates in existing GDRO methods have the same order of computational cost~\citep{zhang2026-GDRO-TPAMI}.
    \item \textbf{$w$-player update.} With Euclidean regularizer, the FTRL update~\eqref{eqn:update:wt} requires $O(d)$ arithmetic operations and one projection onto $\W$, in addition to computing one stochastic gradient, where $d$ is the model dimension. This is comparable to a Euclidean OMD update, which also requires $O(d)$ arithmetic operations and one projection. The stochastic gradient reuses the exploitation sample and thus requires no additional sample query.
    \item \textbf{Group selection and sample acquisition.} Drawing $c_t$ from $\q_t$ and selecting the exploration set $\I_t$ can be implemented in $O(m)$ time, e.g., DepRound in Appendix~\ref{appendix:section:DepRound}. Each round then acquires $r_t$ samples and evaluates their losses, while computing only one stochastic gradient for the $w$-player.
\end{compactenum}

\section{Supporting Algorithm: DepRound} \label{appendix:section:DepRound}
The DepRound algorithm has been extensively employed in combinatorial semi-bandit algorithms \citep{DepRound-app1, Exp4.MP, Online:Bandit:Minimax}.
DepRound takes as input a vector $\p\in\R^m$ satisfying $0\leq p_i\leq 1$ and $\|\p\|_1=k\in\mathbb{Z}^+$, and outputs a set $\I$ such that $\Pr[i\in \I]=p_i$ and $|\I|=k$, with a computational complexity $O(m)$ for both time and space. 
The DepRound procedure is outlined in Algorithm~\ref{alg:DepRound}.
\begin{algorithm}[H]
	\caption{DepRound}
	\label{alg:DepRound}
	\textbf{Input}: vector $\p\in\R^m$ that satisfies $\mathbf{0} \leq\p\leq\mathbf{1} $ and $\|\p\|_1=k\in\mathbb{Z}^+$
	\begin{algorithmic}[1]
		\WHILE{$\exists i\in [m]$ s.t. $p_i\in(0,1)$}
		\STATE Choose any $i,j\in[m]$ with $i\neq j$ and $p_i\in(0,1),p_j\in(0,1)$
		\STATE Set $\alpha=\min\{1-p_i,p_j\}$ and $\beta=\min\{p_i,1-p_j\}$
		\STATE Update 
		$$(p_i,p_j)=\begin{cases}(p_i+\alpha,p_j-\alpha)\text{ with probability }\frac{\beta}{\alpha+\beta}\\(p_i-\beta,p_j+\beta)\text{ with probability }\frac{\alpha}{\alpha+\beta}\end{cases}$$
		\ENDWHILE
		\STATE \textbf{return} $\I=\{i\in[m]\mid p_i=1\}$
	\end{algorithmic}
\end{algorithm}

\section{Analysis}  \label{appendix:Analysis}
In analysis, $\mathcal{F}_{t-1}$ contains the history through round $t-1$ and the currently revealed sample size $r_t$, before the learner's round-$t$ randomization and fresh sampling. In particular, $r_t$, $\w_t$, $\q_t$, $\eta_{q,t}$, $\eta_{w,t}$, and $\gamma_t$ are $\mathcal{F}_{t-1}$-measurable. We write $\E_{t-1}[\cdot]=\E[\cdot\mid\mathcal{F}_{t-1}]$.

\subsection{Supporting Lemmas}
Based on the discussion in Section~\ref{sec:preliminary}, we first state an error-decomposition lemma.
\begin{lemma} \label{lem:General_framework} 
	Under Assumptions~\ref{ass:gen_domain_w}-\ref{ass:gen_loss_convex}, suppose that
	\begin{compactenum}
	\item The $w$-player is equipped with a non-oblivious OCO algorithm $\A_w$, and for any fixed $t\in\N$, with probability at least $1 - \delta$, the regret $\Regw$ is upper bounded by $U_w(t,\delta)$.
	\item The $q$-player is equipped with a non-oblivious PLA algorithm $\A_q$, and for any fixed $t\in\N$, with probability at least $1 - \delta$, the regret $\Regqp$ is upper bounded by $U_q(t,\delta)$.
	\end{compactenum}
Then, for any fixed $t\in\N$, with probability at least $1-\delta$, we have
\begin{equation} \label{eqn:lem:General_framework:result:high-pro-weak}
	\epsilon_{\phi}(\wb_t, \qb_t)
	\leq \frac{1}{t}U_w\left(t,\frac{\delta}{4}\right) + \frac{1}{t}U_q\left(t,\frac{\delta}{4}\right) + 2\sqrt{\frac{2}{t} \ln \frac{4m}{\delta}}.
\end{equation}
\end{lemma}

Then, we give some supporting lemmas as follows. 
\begin{lemma}\citep[Lemma 7]{pmlr-v32-seldin14}\label{lem:sample_bound}
	For any probability distribution $\q\in\Delta_m$ and any $r\in[m]$, we have
	\begin{equation} \label{eqn:lem:inequality_bound}
		\sum_{i=1}^m\frac{q_i(m-1)}{q_i(m-r)+r-1}\leq\frac mr.
	\end{equation}
\end{lemma}

\begin{lemma}\citep[Lemma 8]{pmlr-v32-seldin14}  \label{lem:varying_stepsize}
	For any sample-size process of non-negative numbers $a_1,a_2,\ldots, a_t$, such that $a_1>0$, and any power $\gamma\in(0,1)$ we have
	\begin{equation} \label{eqn:lem:varying_stepsize}	\sum_{j=1}^t\frac{a_j}{\left(\sum_{s=1}^ja_s\right)^\gamma}\leq\frac1{1-\gamma}\left(\sum_{j=1}^ta_j\right)^{1-\gamma}.
	\end{equation}
\end{lemma}

We present Azuma's inequality and an adaptive-budget Bernstein inequality as follows.
\begin{lemma} \citep[Lemma A.7]{cesa2006prediction}  \label{lem:azuma}
	Let $V_1, V_2,  \ldots$  be a martingale difference sequence with respect to some sequence $X_1, X_2, \ldots$ such that $V_i \in [A_i , A_i + c_i ]$ for some random variable $A_i$, measurable with respect to $X_1, \ldots , X_{i-1}$ and a positive constant $c_i$. If $S_n = \sum_{i=1}^n V_i$, then for any
	$t > 0$,
	\[
	\Pr[ S_n > t] \leq \exp \left( -\frac{2t^2}{\sum_{i=1}^n c_i^2} \right).
	\]
\end{lemma}
\begin{lemma}\label{lem:Bernstein_inequality}
    Fix $t\in\N$, an integer $m\geq2$, and $\delta\in(0,1)$. Let $\{X_j\}_{j=1}^t$ be a martingale difference sequence with $|X_j|\leq b$ for $b>0$. Let $\{r_j\}_{j=1}^t$ follow Definition~\ref{def:sample_size_protocol}. Suppose a fixed nonnegative and nondecreasing function $\Psi:(0,\infty)\to[0,\infty)$ satisfies, for every $B>0$, $\sum_{j=1}^t\E_{j-1}\big[(X_j\ind\left[\sum_{s=1}^j\frac{m}{r_s}\leq B\right])^2\big]\leq\Psi(B)$. 
    Then, with probability at least $1-\delta$,
    \begin{equation}\label{eqn:adaptive_budget_Bernstein}
        \sum_{j=1}^t X_j\leq\sqrt{2\Psi\left(2\sum_{j=1}^t\frac{m}{r_j}\right)\ln\frac{m}{\delta}}+\frac{2b}{3}\ln\frac{m}{\delta}.
    \end{equation}
\end{lemma}
The proof is given in Appendix~\ref{sec:proof_Bernstein_inequality}.

Note that both Lemmas~\ref{lem:azuma} and~\ref{lem:Bernstein_inequality} hold only for a fixed time horizon $t$. In the following, we present two time-uniform versions to establish a time-uniform guarantee.
\begin{lemma} \citep[Lemma 27]{lee2025lasso}    \label{lem:azuma_time_uniform}
Let $\{X_t\}_{t=1}^\infty$ be a real-valued martingale difference sequence adapted to a filtration $\{\mathcal{F}_t\}_{t=0}^\infty$. Assume that $\{X_t\}_{t=1}^\infty$ is conditionally $\sigma$-sub-Gaussian, i.e., $\E[e^{sX_t}\mid \mathcal{F}_{t-1}] \leq e^{\frac{s^2\sigma ^2}{2}}$ for all $s\in \R$. Then, the following inequality holds with probability at least $1-\delta$, for all $n\in\N$:
\begin{equation*}
	\sum_{t=1}^nX_t\leq2^{\frac34}\sigma\sqrt{n\ln\frac{7(\ln2n)^2}{2\delta}}.
\end{equation*}
\end{lemma}
Although a time-uniform version of Bernstein’s inequality is presented by \citet[Lemma 37]{lee2025minimax}, we adopt the Theorem~3 from the author's blog \footnote{\url{https://harinboy.github.io/posts/FreedmansInequality/}} as it provides a simpler bound. For completeness, we provide its proof in Appendix~\ref{sec:proof_Bernstein_inequality_time_uniform}.
\begin{lemma}  \label{lem:Bernstein_inequality_time_uniform}
Let $\{X_n\}_{n=1}^\infty$ be a martingale difference sequence with respect to a filtration $\{\mathcal{F}_n\}_{n=0}^\infty$. Suppose $X_n \leq 1$ holds almost surely for all $n$. Let $T_n = \sum_{t=1}^n \mathbb{E}[X_t^2|\mathcal{F}_{t-1}]$. Then, the following inequality holds with probability at least $1-\delta$, for all $n\in\N$:
\begin{equation*}
	\sum_{i=1}^n X_i \leq 2 \ln \frac{2(1+\ln^+ T_n)^2}{\delta} + 2\sqrt{T_n \ln \frac{2(1+\ln^+ T_n)^2}{\delta}}
\end{equation*}
where $\ln^+ x = \max\{\ln x, 0\}$ for $x>0$, with the convention $\ln^+0=0$.
\end{lemma}

\subsection{Proof of Lemma~\ref{lem:GDRO_q:col}} \label{sec:proof_GDRO_q}
We first present the complete form of Lemma~\ref{lem:GDRO_q:col}.
\begin{lemma} \label{lem:GDRO_q:col_formal} 
	Let $\eta_{q,t}$ and $\gamma_t$ be defined in \eqref{eqn:proof:q:stepsize:1} and~\eqref{eqn:proof:q:stepsize:sep:gammat} for Algorithm~\ref{alg:q_GDRO:col}. Under Assumption~\ref{ass:gen_loss_bounded}, we have for any fixed $t\in\N$ and any sample-size process $\{r_j\}_{j\in[t]}$ generated according to Definition~\ref{def:sample_size_protocol}, with probability at least $1-\delta$,
\begin{equation*}\label{eqn:lem:GDRO_q:High-pro:col_formal}
	\begin{split}
\Regqp\leq{}&
\sqrt{\sum_{j=1}^t\frac{m}{r_j}}
\left(9\sqrt{2\ln m}+4\sqrt{2\ln\frac{5}{\delta}}\right)+2^{1/4}m\sqrt{\ln m\sqrt{\sum_{j=1}^t\frac{m}{r_j}}\ln\frac{5m}{\delta}}\\
&+\left(m+1+\frac{m^2\sqrt{\ln m}}{3}+\sqrt{\frac{2\sum_{j=1}^t\frac{m}{r_j}}{\ln m}}
\right)\ln\frac{5}{\delta}+\left(\frac{m^2\sqrt{\ln m}}{3}+\frac{5m}{3}\right)\ln m.
	\end{split}
\end{equation*}
\end{lemma}

By defining $\nu_{q,t}(\q)=\frac{1}{\eta_{q,t}}\left(\nu_q(\q)-\min_{\q \in \Delta_m}\nu_q(\q)\right)$ with the negative-entropy regularizer $\nu_q(\q) = \sum_{i=1}^{m} q_i \log q_i$, the update rule in \eqref{eqn:GDRO:qt_update:col} can be reformulated as the standard FTRL update rule
\begin{equation*}
	\q_{t}=\argmin_{\q\in\Delta_m}\left\{\inner{\Lb_{t-1}}{\q}+\nu_{q,t}(\q)\right\}.
\end{equation*}
Since $\eta_{q,t}$ is a non-increasing sequence, by the results from \citet[Remark 7.4., Lemma 7.14 and \S 7.5]{orabona2023}, we can derive for each $i\in[m]$,
\begin{equation}  \label{eqn:proof:FTRL_local_norm:GDRO}
	\sum_{j=1}^t \left(\inner{\q_j}{\tildesb_j}-{\tildes_{j,i}}\right)
	\leq\frac{\ln m}{\eta_{q,t}}+	\sum_{j=1}^t\frac{\eta_{q,j}}{2}\sum_{i=1}^{m}q_{j,i}(\tildes_{j,i})^{2}.
\end{equation}

Using the notation $k^* \in \argmin_{i \in [m]} \sum_{j=1}^t \hats_{j,i}$, $S_t=\{j|r_j=1,j\in[t]\}$ and $ M_t=\{j|r_j\geq2,j\in[t]\}$, we have
\begin{equation} \label{eq:proof:q:tl:1}
	\begin{split}
		\Regqp=&\sum_{j=1}^t \inner{\q_j}{\hatsb_j} - \min_{i\in[m]} \sum_{j=1}^t \hats_{j,i} = \sum_{j=1}^t \inner{\q_j}{\hatsb_j} - \sum_{j=1}^t \hats_{j,k^*}\\
		=&\sum_{j=1}^t \left(\inner{\q_j}{\tildesb_j}-{\tildes_{j,k^*}}\right)+\sum_{j=1}^t \left(\inner{\q_j}{\hatsb_j}- \inner{\q_j}{\tildesb_j}\right)+\sum_{j=1}^t\left(\tildes_{j,k^*} -  \hats_{j,k^*}\right)\\
		\overset{\eqref{eqn:proof:FTRL_local_norm:GDRO}}{\leq}&\frac{\ln m}{\eta_{q,t}}+	\sum_{j=1}^t\frac{\eta_{q,j}}{2}\sum_{i=1}^{m}q_{j,i}(\tildes_{j,i})^{2}+\sum_{j=1}^t \left(\inner{\q_j}{\hatsb_j}- \inner{\q_j}{\tildesb_j}\right)+\sum_{j=1}^t\left(\tildes_{j,k^*} -  \hats_{j,k^*}\right).
	\end{split}
\end{equation}
Then, we analyze the bias introduced by $\tilde{s}_{j,i}$ when $j\in S_t$.
For $r_j=1$, the selected distribution set $C_j=\{c_j\}$. Using \citet[(5)]{Exp3-IX}, for $j\in S_t$, we have
\begin{equation} \label{eq:proof:q:Es:0}
	\inner{\q_j}{\tildesb_j}=\sum_{i=1}^m\ind\left[i=c_j\right]\frac{\hats_{j,i}\left(q_{j,i}+\gamma_j\right)}{q_{j,i}+\gamma_j}-\gamma_j\sum_{i=1}^m\ind\left[i= c_j\right]\frac{\hats_{j,i}}{q_{j,i}+\gamma_j}=\hats_{j,c_j}-\gamma_j\sum_{i=1}^m\tildes_{j,i}.
\end{equation}
By combining \eqref{eq:proof:q:Es:0} with \eqref{eq:proof:q:tl:1}, and noting that $S_t\cup M_t=[t]$ and $q_{j,i}\tildes_{j,i}\leq1$ for all $j\in S_t, i\in[m]$, we obtain
\begin{equation} \label{eq:proof:q:tl:2}
	\begin{split}
		\Regqp
		\leq&\frac{\ln m}{\eta_{q,t}}+\sum_{j\in S_t}\frac{\eta_{q,j}}{2}\sum_{i=1}^{m}\tildes_{j,i}+\sum_{j\in M_t}\frac{\eta_{q,j}}{2}\sum_{i=1}^{m}q_{j,i}(\tildes_{j,i})^{2}\\
		&+\sum_{j\in S_t} \left(\inner{\q_j}{\hatsb_j}- \inner{\q_j}{\tildesb_j}\right)+\sum_{j\in M_t} \left(\inner{\q_j}{\hatsb_j}- \inner{\q_j}{\tildesb_j}\right)+\sum_{j=1}^t\left(\tildes_{j,k^*} -  \hats_{j,k^*}\right)\\
		\overset{\eqref{eq:proof:q:Es:0}}{\leq}&\frac{\ln m}{\eta_{q,t}}+\underbrace{\sum_{j\in S_t}\left(\frac{\eta_{q,j}}{2}+\gamma_j\right)\sum_{i=1}^{m}\tildes_{j,i}}_{\mathtt{term (d)}}+\underbrace{\sum_{j\in M_t}\frac{\eta_{q,j}}{2}\sum_{i=1}^{m}q_{j,i}(\tildes_{j,i})^{2}}_{\mathtt{term (e)}}\\
		&+\underbrace{\sum_{j\in S_t} \left(\inner{\q_j}{\hatsb_j}- \hats_{j,c_j}\right)+\sum_{j\in M_t} \left(\inner{\q_j}{\hatsb_j}- \inner{\q_j}{\tildesb_j}\right)}_{\mathtt{term (f)}}\\
		&+\underbrace{\sum_{j\in S_t}\left(\tildes_{j,k^*} -  \hats_{j,k^*}\right)}_{\mathtt{term (g)}}+\underbrace{\sum_{j\in M_t}\left(\tildes_{j,k^*} -\hats_{j,k^*}\right)}_{\mathtt{term (h)}}.
	\end{split}
\end{equation}

For $\mathtt{term (d)}$, we invoke Lemma~2 in \citet{zhang2023-SA-GDRO} and extend it to any subset of $[t]$ as follows.
\begin{lemma} \label{lem:high-prob-martingale:Exp3-IX}
	Let $\mathcal{F}_{j-1}$ denote the analysis history before round-$j$ sampling, including all past loss vectors and the currently revealed sample size. Let $\p_j\in\Delta_m$ be $\mathcal{F}_{j-1}$-measurable, and let $\hat{\xi}_{j,i}\in[0,1]$ with $\Pr(i_j=i\mid\mathcal{F}_{j-1},\hat{\xi}_{j,1},\ldots,\hat{\xi}_{j,m})=p_{j,i}$. Define the IX estimator $\tilde{\xi}_{j,i}=\frac{\hat{\xi}_{j,i}}{p_{j,i}+\gamma_j}\indicator{i_j=i}$. Let $\{\gamma_j\}_{j=1}^t$ be positive and non-increasing, with $\gamma_j$ being $\mathcal{F}_{j-1}$-measurable for every $j\in[t]$, and let $\alpha_{j,i}$ be non-negative $\mathcal{F}_{j-1}$-measurable random variables satisfying $\alpha_{j,i}\leq2\gamma_j$. Let $S\subseteq[t]$ have $\mathcal{F}_{j-1}$-measurable indicators $\ind[j\in S]$. Then, with probability at least $1-\delta$,
	\begin{equation} \label{eqn:high-prob-margingale}
		\sum_{j\in S} \sum_{i=1}^m \alpha_{j,i} (\tilde{\xi}_{j,i} - \hat{\xi}_{j,i}) \leq \ln \frac{1}{\delta}.
	\end{equation}
\end{lemma}
For our application, the full vector of fresh losses can be generated independently of the current index draw, conditional on the past and $r_j$, with only the queried coordinates revealed. Moreover, $r_j$ is $\mathcal{F}_{j-1}$-measurable by Definition~\ref{def:sample_size_protocol}, so $\ind[r_j=1]$ is $\mathcal{F}_{j-1}$-measurable; $\gamma_j$ and $\eta_{q,j}$ are also $\mathcal{F}_{j-1}$-measurable by~\eqref{eqn:proof:q:stepsize:1} and~\eqref{eqn:proof:q:stepsize:sep:gammat}. Thus, for $j\in S_t$, our construction satisfies the requirements of Lemma~\ref{lem:high-prob-martingale:Exp3-IX}. Setting $S=S_t$ and $\alpha_{j,i}=\eta_{q,j}$ for all $j\in[t]$ and $i\in[m]$, we have, with probability at least $1-\delta$,
\begin{equation}\label{eq:proof:q:A}
	\begin{split}
		&\sum_{j\in S_t}\left(\frac{\eta_{q,j}}{2}+\gamma_j\right)\sum_{i=1}^{m}\tildes_{j,i}
		\overset{\eqref{eqn:proof:q:stepsize:sep:gammat}}{=} \sum_{j\in S_t}\eta_{q,j}\sum_{i=1}^{m}\tildes_{j,i}
		\overset{\eqref{eqn:high-prob-margingale}}{\leq} \sum_{j\in S_t}\eta_{q,j}\sum_{i=1}^{m}\hats_{j,i}+\ln \frac{1}{\delta}
		\overset{\eqref{def:hats:main_part}}{\leq}  \sum_{j\in S_t}\eta_{q,j}m+\ln \frac{1}{\delta} \\
		\overset{\eqref{eqn:proof:q:stepsize:1}}{=} & \sum_{j\in S_t}\sqrt{\frac{m\ln m}{\sum_{s=1}^j\frac{1}{r_s}}}+\ln \frac{1}{\delta}
		\leq \sum_{j\in S_t} \sqrt{\frac{m\ln m}{|[j]\cap S_t|}}+\ln \frac{1}{\delta} 
		\leq   2\sqrt{|S_t|m\ln m}+\ln \frac{1}{\delta},
	\end{split}
\end{equation}
where the last inequality follows from the fact $\sum_{j=1}^t\frac{1}{\sqrt{j}}\leq\int_0^{t}\frac{1}{\sqrt{x}}\text{d}x=2\sqrt{x}|_0^{t}=2\sqrt{t}$.

For $\mathtt{term (e)}$, we denote $Q_j=\frac{\eta_{q,j}}{2}\sum_{i=1}^{m}q_{j,i}(\tildes_{j,i})^{2}$ for $j\in [t]$. 
For clarity and ease of reference, we present the following lemma.
\begin{lemma}\label{lem:Bernstein_inequality_extension}
Under Assumption~\ref{ass:gen_loss_bounded} and the sampling rule of Algorithm~\ref{alg:q_GDRO:col} with the sample-size protocol in Definition~\ref{def:sample_size_protocol}, let $\eta_{q,j}$ be defined in~\eqref{eqn:proof:q:stepsize:1}, $Q_j=\frac{\eta_{q,j}}{2}\sum_{i=1}^m q_{j,i}\tildes_{j,i}^2$, and $M_t=\{j\in[t]:r_j\geq2\}$. Then, for any fixed $t\in\mathbb{N}_+$ and $\delta\in(0,1)$, with probability at least $1-\delta$,
\begin{equation}\label{eqn:lem:Bernstein_inequality_extension}
	\sum_{j\in M_t}Q_j\leq 2^{1/4}m\sqrt{\ln m\sqrt{\sum_{j=1}^t\frac{m}{r_j}}\ln\frac{m}{\delta}}+\frac{m^2\sqrt{\ln m}}{3}\ln\frac{m}{\delta}+\frac{m}{2}\sum_{j\in M_t}\frac{\eta_{q,j}}{r_j}.
\end{equation}
\end{lemma}
By setting $\gamma=\frac{1}{2}$ in Lemma~\ref{lem:varying_stepsize} and applying it to the subsequence indexed by $M_t$, we have
\begin{equation}\label{eq:proof:q:B_2}
\begin{split}
\frac{m}{2}\sum_{j\in M_t}\frac{\eta_{q,j}}{r_j}
\overset{\eqref{eqn:proof:q:stepsize:1}}{=}&\frac{\sqrt{m\ln m}}{2}\sum_{j\in M_t}\frac{\frac{1}{r_j}}{\sqrt{\sum_{s=1}^j\frac{1}{r_s}}}\\
\leq&\frac{\sqrt{m\ln m}}{2}\sum_{j\in M_t}\frac{\frac{1}{r_j}}{\sqrt{\sum_{s\in[j]\cap M_t}\frac{1}{r_s}}}
\overset{\eqref{eqn:lem:varying_stepsize}}{\leq}\sqrt{\sum_{j\in M_t}\frac{m}{r_j}\ln m}.
\end{split}
\end{equation}
Combining Lemma~\ref{lem:Bernstein_inequality_extension} with~\eqref{eq:proof:q:B_2}, we obtain, with probability at least $1-\delta$,
\begin{equation}\label{eq:proof:q:B}
\sum_{j\in M_t}Q_j\leq 2^{1/4}m\sqrt{\ln m\sqrt{\sum_{j=1}^t\frac{m}{r_j}}\ln\frac{m}{\delta}}+\frac{m^2\sqrt{\ln m}}{3}\ln\frac{m}{\delta}+\sqrt{\sum_{j\in M_t}\frac{m}{r_j}\ln m}.
\end{equation}

For $\mathtt{term (f)}$, we define $V_j=\inner{\q_j}{\hatsb_j}-\hats_{j,c_j}$ when $j\in S_t$ and $V_j=\inner{\q_j}{\hatsb_j}-\inner{\q_j}{\tildesb_j}$ when $j\in M_t$. Then we know $\E_{j-1}\left[V_j\right]=0$. By Lemma~\ref{lem:sample_bound}, \eqref{eqn:def:sti}, \eqref{def:hats:main_part}, $m\geq2$ and $r_j\geq2$ when $j\in M_t$, we know $|V_j|\leq \frac{m}{2}$ for all $j\in[t]$. Moreover, for $j\in S_t$, we have
\begin{equation*}
	\E_{j-1}\left[V_j^2\right]
	=\E_{j-1}\left[\left(\inner{\q_j}{\hatsb_j}- \hats_{j,c_j}\right)^2\right]\leq 1,
\end{equation*}
and for $j\in M_t$, we have
\begin{align*}
	\E_{j-1}\left[V_j^2\right]
	&=\E_{j-1}\left[\left(\inner{\q_j}{\hatsb_j}-\inner{\q_j}{\tildesb_j}\right)^2\right]
	=\E_{j-1}\left[\left(\inner{\q_j}{\tildesb_j}\right)^2\right]-\E_{j-1}\left[\left(\inner{\q_j}{\hatsb_j}\right)^2\right]\\
	&\leq \E_{j-1}\left[\left(\inner{\q_j}{\tildesb_j}\right)^2\right]
	\overset{\eqref{eqn:lem:inequality_bound}}{\leq}\frac{m}{r_j}\E_{j-1}\left[\inner{\q_j}{\tildesb_j}\right]\leq \frac{m}{r_j}.
\end{align*}
In both regimes, $\E_{j-1}[V_j^2]\leq \frac{m}{r_j}$. To apply Lemma~\ref{lem:Bernstein_inequality}, we instantiate its martingale difference sequence as $X_j=V_j$. Let $A_j=\sum_{s=1}^j\frac{m}{r_s}$. For any $B>0$, since $\ind[A_j\leq B]$ is $\mathcal{F}_{j-1}$-measurable, we have
\[
\sum_{j=1}^t\E_{j-1}\left[\left(V_j\ind[A_j\leq B]\right)^2\right]=\sum_{j=1}^t\ind[A_j\leq B]\E_{j-1}[V_j^2]\leq\sum_{j=1}^t\frac{m}{r_j}\ind[A_j\leq B]\leq B.
\]
Therefore, the conditions of~\eqref{eqn:adaptive_budget_Bernstein} hold with $\Psi(B)=B$ and $b=m/2$. It follows that, with probability at least $1-\delta$,
\begin{equation}\label{eq:proof:q:C}
\sum_{j=1}^t V_j\leq 2\sqrt{\sum_{j=1}^t\frac{m}{r_j}\ln\frac{m}{\delta}}+\frac{m}{3}\ln\frac{m}{\delta}.
\end{equation}

For $\mathtt{term (g)}$, we need to control the unweighted sum $\sum_{j\in S_t}(\tildes_{j,i}-\hats_{j,i})$ for each $i\in[m]$. Lemma~\ref{lem:high-prob-martingale:Exp3-IX}, however, controls weighted sums with weights bounded by $2\gamma_j$. To obtain the desired unweighted bound, we apply this lemma with a family of scale-dependent common weights. Let $A_j=\sum_{s=1}^j\frac{m}{r_s}$. Fix $t$, let $K=1+\lfloor\log_2m\rfloor\leq m$, and set $B_k=2^{k+1}t$ and $\lambda_k=\sqrt{\ln(m)/B_k}$ for $k=0,\ldots,K-1$. For each $i\in[m]$, define $\alpha_{j,\ell}^{(k,i)}=\lambda_k\ind[A_j\leq B_k]\ind[\ell=i]$. Since $A_j$ is $\mathcal{F}_{j-1}$-measurable and $\lambda_k\leq\sqrt{\ln(m)/A_j}=2\gamma_j$ whenever $A_j\leq B_k$, these weights satisfy the conditions of Lemma~\ref{lem:high-prob-martingale:Exp3-IX}. Applying the lemma with $S=S_t$ and failure probability $\delta/(mK)$, followed by a union bound over $i$ and $k$, gives, with probability at least $1-\delta$,
\[
\lambda_k\sum_{j\in S_t}\ind[A_j\leq B_k](\tildes_{j,i}-\hats_{j,i})\leq\ln\frac{mK}{\delta}\leq\ln\frac{m^2}{\delta},
\]
simultaneously for all $i\in[m]$ and $k=0,\ldots,K-1$.
Since $t\leq A_t\leq mt$, there exists $k\in\{0,\ldots,K-1\}$ such that $2^kt\leq A_t<B_k\leq2A_t$. For this $k$, $A_j\leq A_t<B_k$ for every $j\leq t$, so all indicators above equal one. Moreover, $\lambda_k^{-1}=\sqrt{B_k/\ln m}\leq\sqrt{2A_t/\ln m}$. Therefore, simultaneously for all $i\in[m]$,
\begin{equation}\label{eq:proof:q:D}
\sum_{j\in S_t}(\tildes_{j,i}-\hats_{j,i})\leq\sqrt{\frac{2A_t}{\ln m}}\ln\frac{m^2}{\delta}=2\sqrt{2\sum_{j=1}^t\frac{m}{r_j}\ln m}+\sqrt{\frac{2\sum_{j=1}^t\frac{m}{r_j}}{\ln m}}\ln\frac{1}{\delta}.
\end{equation}

For $\mathtt{term (h)}$, define $X_{j,i}=(\tildes_{j,i}-\hats_{j,i})\ind[r_j\geq2]$ for all $j\in[t]$ and $i\in[m]$. Then $\E_{j-1}[X_{j,i}]=0$ and $|X_{j,i}|\leq m$. For $j\in M_t$,
\[
\E_{j-1}[X_{j,i}^2]\leq\E_{j-1}[\tildes_{j,i}^2]\leq\frac{1}{q_{j,i}+(1-q_{j,i})\frac{r_j-1}{m-1}}\leq\frac{m}{r_j-1}\leq\frac{2m}{r_j}.
\]
Hence, for any $B>0$,
\[
\sum_{j=1}^t\E_{j-1}\left[\left(X_{j,i}\ind[A_j\leq B]\right)^2\right]\leq2\sum_{j=1}^t\frac{m}{r_j}\ind[A_j\leq B]\leq2B.
\]
Applying~\eqref{eqn:adaptive_budget_Bernstein} with $\Psi(B)=2B$, $b=m$, and failure probability $\delta/m$, and then taking a union bound over $i\in[m]$, we obtain, with probability at least $1-\delta$, simultaneously for all $i\in[m]$,
\begin{equation}\label{eq:proof:q:E}
\sum_{j\in M_t}(\tildes_{j,i}-\hats_{j,i})\leq2\sqrt{2\sum_{j=1}^t\frac{m}{r_j}\ln\frac{m^2}{\delta}}+\frac{2m}{3}\ln\frac{m^2}{\delta}.
\end{equation}

Combining~\eqref{eq:proof:q:A}--\eqref{eq:proof:q:E} with~\eqref{eq:proof:q:tl:2} and allocating failure probability $\delta/5$ to each term, and further using~\eqref{eqn:proof:q:stepsize:1}, $\ln(5m/\delta)=\ln m+\ln(5/\delta)$, $\ln(5m^2/\delta)=2\ln m+\ln(5/\delta)$, and $\sqrt{a+b}\leq\sqrt{a}+\sqrt{b}$, we obtain, with probability at least $1-\delta$,
\begin{equation*}
\begin{split}
\Regqp\leq{}&
\sqrt{\sum_{j=1}^t\frac{m}{r_j}}
\left(9\sqrt{2\ln m}+4\sqrt{2\ln\frac{5}{\delta}}\right)+2^{1/4}m\sqrt{\ln m\sqrt{\sum_{j=1}^t\frac{m}{r_j}}\ln\frac{5m}{\delta}}\\
&+\left(m+1+\frac{m^2\sqrt{\ln m}}{3}+\sqrt{\frac{2\sum_{j=1}^t\frac{m}{r_j}}{\ln m}}
\right)\ln\frac{5}{\delta}+\left(\frac{m^2\sqrt{\ln m}}{3}+\frac{5m}{3}\right)\ln m.
\end{split}
\end{equation*}
Here, we also use $|S_t|m\leq\sum_{j=1}^t\frac{m}{r_j}$ and
\begin{equation}\label{eqn:proof:result_scale}
\sqrt{\sum_{j\in M_t}\frac{m}{r_j}\ln m}+\sqrt{|S_t|m\ln m}\leq\sqrt{2\sum_{j=1}^t\frac{m}{r_j}\ln m},
\end{equation}
where the latter follows from $S_t\cup M_t=[t]$ and $r_j=1$ for $j\in S_t$.

\subsection{Proof of Lemma~\ref{lem:wt}}\label{sec:proof_xt}
By convexity of $\phi(\cdot,q_j)$, we have
\begin{equation}  \label{eqn:lem:proof:xt:decom}
	\begin{split}
		&\sum_{j=1}^t \left[ \phi(\w_j,\q_j) -  \phi(\w,\q_j) \right] \leq  \sum_{j=1}^t \langle \nabla_\w \phi(\w_j,\q_j), \w_j-\w\rangle\\
		\leq& \sum_{j=1}^t\langle \g_{w}(\w_j,\q_j), \w_j-\w\rangle + \sum_{j=1}^t \langle \nabla_\w \phi(\w_j,\q_j) - \g_{w}(\w_j,\q_j), \w_j-\w\rangle.
	\end{split}
\end{equation}

We bound the first term as follows. By defining $\nu_{w,t}\left(\w\right)=\frac{1}{\eta_{w,t}}\left(\nu_w(\w)-\min_{\w \in \W}\nu_w(\w)\right)$, we rewrite \eqref{eqn:update:wt} as 
\begin{equation*}\label{eqn:proof:upate:xt}
	\w_t= \argmin_{\w \in \W} \left\{ \left\langle\sum_{i=1}^{t-1}\g_w(\w_i,\q_i),\w\right\rangle + \nu_{w,t}\left(\w\right) \right\}.
\end{equation*}
By the standard analysis of FTRL~\citep[Corollary~7.9.]{orabona2023} and the fact that $\nu_w$ is $1$-strongly convex with respect to a certain norm $\|\cdot\|_w$, for all $t\in\mathbb{Z}^+$ and $\w\in\W$, we have
\begin{equation} \label{eqn:FTRL:0}
	\begin{split}
		\sum_{j=1}^t \langle  \g_{w}(\w_j,\q_j), \w_j - \w \rangle 
		\leq &\frac{\nu_w(\w) - \min_{\w\in \W}\nu_w(\w) }{\eta_{w,t}} + \frac{1}{2}\sum_{j=1}^t\eta_{w,j} \| \g_{w}(\w_j,\q_j)\|_{w,*}^2\\
		\overset{\eqref{eqn:ass:domain:W},\eqref{eqn:update:wt:unbiased_gradient}}{\leq} & \frac{D^2}{\eta_{w,t}} + \frac{G^2}{2}\sum_{j=1}^{t}\eta_{w,j}.
	\end{split}
\end{equation}

Taking the maximum over the two sides in \eqref{eqn:lem:proof:xt:decom} over $\w \in \W$, we obtain
\begin{equation} \label{eqn:lem:proof:xt:decom:2}
	\begin{split}
		&\max_{\w\in\W}  \left\{\sum_{j=1}^t \left[ \phi(\w_j,\q_j) -  \phi(\w,\q_j) \right]\right\}= \sum_{j=1}^t \phi(\w_j,\q_j) - \min_{\w\in\W} \sum_{j=1}^t \phi(\w,\q_j)\\
		\overset{\eqref{eqn:FTRL:0}}{\leq} &  \frac{D^2}{\eta_{w,t}} + \frac{G^2}{2}\sum_{j=1}^{t}\eta_{w,j} +\max_{\w \in \W} \left\{\underbrace{\sum_{j=1}^t \langle \nabla_\w \phi(\w_j,\q_j) - \g_{w}(\w_j,\q_j), \w_j-\w\rangle}_{=F_t(\w)} \right\}.
	\end{split}
\end{equation}

To bound $\max_{\w \in \W} F_t(\w)$ in \eqref{eqn:lem:proof:xt:decom:2}, we cannot directly treat the summands of $F_t(\wt)$ as martingale differences, because $\wt = \argmax_{\w \in \W} F_t(\w)$ may depend on the gradient noise from rounds $1,\ldots,t$ \citep{zhang2023-SA-GDRO}. To address this challenge, we employ the ``ghost iterate'' technique \citep[proof of Lemma 3.1]{nemirovski-2008-robust}, which decouples the dependency between $\wt$ and $F_t(\w)$. It is important to note that the referenced work applies the ``ghost iterate'' to SMD, whereas our analysis constructs an FTRL-based variant. Specifically, we introduce a virtual sequence of variables that performs FTRL as
\begin{equation} \label{eqn:virtual:FTRL}
	\vv_t= \argmin_{\w \in \W} \left\{ \left\langle\sum_{j=1}^{t-1} \nabla_\w \phi(\w_j,\q_j) - \g_{w}(\w_j,\q_j) , \w\right\rangle + \nu_{w,t}\left(\w\right) \right\},
\end{equation}
where $\vv_1=\w_1$. By repeating the derivation of \eqref{eqn:FTRL:0}, we obtain for all $\w\in\W$,
\begin{equation} \label{eqn:virtual:FTRL:res}
	\begin{split}
		&\sum_{j=1}^t \langle  \nabla_\w \phi(\w_j,\q_j) - \g_{w}(\w_j,\q_j), \vv_j - \w \rangle  \\
		\leq  & \frac{\nu_w(\w) - \min_{\w\in \W}\nu_w(\w) }{\eta_{w,t}} + \frac{1}{2}\sum_{j=1}^t\eta_{w,j} \| \nabla_\w \phi(\w_j,\q_j) - \g_{w}(\w_j,\q_j)\|_{w,*}^2 \\
		\leq & \frac{D^2}{\eta_{w,t}} + 2G^2\sum_{j=1}^{t}\eta_{w,j},
	\end{split}
\end{equation}
where the last inequality is due to \eqref{eqn:ass:domain:W} and
\begin{equation} \label{eqn:smd:5}
	\begin{split}
		&\| \nabla_\w \phi(\w_j,\q_j) - \g_{w}(\w_j,\q_j)\|_{w,*} \leq \|\nabla_\w\phi(\w_j,\q_j)\|_{w,*} + \|\g_{w}(\w_j,\q_j)\|_{w,*}  \\
		\leq &\E_{j-1}[\| \g_{w}(\w_j,\q_j)\|_{w,*}] + \|\g_{w}(\w_j,\q_j)\|_{w,*}  \overset{\eqref{eqn:update:wt:unbiased_gradient}}{\leq} 2G.
	\end{split}
\end{equation}
Then, we have
\begin{equation} \label{eqn:smd:6}
	\begin{split}
		\max_{\w \in \W} F_t(\w)
		= & \max_{\w \in \W} \left\{\sum_{j=1}^t \langle \nabla_\w \phi(\w_j,\q_j) - \g_{w}(\w_j,\q_j), \vv_j-\w\rangle\right\}  \\
		&+ \sum_{j=1}^t \langle \nabla_\w \phi(\w_j,\q_j) - \g_{w}(\w_j,\q_j), \w_j-\vv_j\rangle   \\
		\overset{\eqref{eqn:virtual:FTRL:res}}{\leq} & \frac{D^2}{\eta_{w,t}} + 2G^2\sum_{j=1}^{t}\eta_{w,j} + \sum_{j=1}^t V_j,
	\end{split}
\end{equation}
where $V_j=\langle \nabla_\w \phi(\w_j,\q_j) - \g_{w}(\w_j,\q_j), \w_j-\vv_j\rangle$.
Since $\vv_j$ and $\w_j$ are $\mathcal{F}_{j-1}$-measurable by~\eqref{eqn:virtual:FTRL} and~\eqref{eqn:update:wt}, and $\E_{j-1}[\g_w(\w_j,\q_j)]=\nabla_\w\phi(\w_j,\q_j)$ by~\eqref{eqn:update:wt:unbiased_gradient}, we have $\E_{j-1}[V_j]=0$ for every $j\in[t]$. Hence $\{V_j\}_{j=1}^t$ is a martingale difference sequence.

To establish a high-probability bound for any fixed $t\in\N$, we first note that
\begin{equation} \label{eqn:smd:upper_bound_Vt}
	\begin{split}
		&|V_j|=\left|\langle \nabla_\w \phi(\w_j,\q_j) - \g_{w}(\w_j,\q_j), \w_j-\vv_j\rangle \right| \\
		\leq &\|\nabla_\w \phi(\w_j,\q_j) - \g_{w}(\w_j,\q_j)\|_{w,*} \|\w_j-\vv_j\|_w \\
		\overset{\text{(\ref{eqn:smd:5})}}{\leq} & 2G \|\w_j-\vv_j\|_w \leq 2G \left( \|\w_j-\w_1\|_w + \|\vv_j-\w_1\|_w \right)\\
		\leq &2G \left( \sqrt{2 B_w(\w_j, \w_1)} + \sqrt{2 B_w(\vv_j, \w_1)} \right) \leq 4 \sqrt{2} D G,
	\end{split}
\end{equation}
where the last step is by defining
\[
B_w(\u,\vv)= \nu_w(\u) - \big[\nu_w(\vv) + \langle \nabla \nu_w(\vv) , \u-\vv\rangle\big], \  \forall \u,\vv\in\W,
\]
and the inequality \citep[(2.42)]{nemirovski-2008-robust}
\begin{equation*} \label{eqn:smd:2}
	\max_{\w \in \W} B_w(\w, \w_1) \leq \max_{\w \in \W} \nu_w(\w) -\min_{\w \in \W} \nu_w(\w)  \overset{\eqref{eqn:ass:domain:W}}{\leq} D^2.
\end{equation*}
By applying Lemma~\ref{lem:azuma}, we obtain for any fixed $t\in\N$, with probability at least $1-\delta$,
\begin{equation} \label{eqn:smd:8}
	\sum_{j=1}^t V_j \leq 8D G \sqrt{t \ln \frac{1}{\delta}}.
\end{equation}
Combining \eqref{eqn:lem:proof:xt:decom:2}, \eqref{eqn:smd:6} and \eqref{eqn:smd:8}, it follows that, with probability at least $1-\delta$,
\begin{equation*}
	\begin{split}
		\sum_{j=1}^t \phi(\w_j,\q_j) - \min_{\w\in\W} \sum_{j=1}^t \phi(\w,\q_j)
		\leq &\frac{2D^2}{\eta_{w,t}} + \frac{5G^2}{2}\sum_{j=1}^{t}\eta_{w,j}+8DG\sqrt{t \ln \frac{1}{\delta}}\\
	   \leq & DG\sqrt{t}\left(2\sqrt{10}+8\sqrt{\ln \frac{1}{\delta}}\right),
	\end{split}
\end{equation*}
where in the last inequality, we set $\eta_{w,j}=\frac{\sqrt{2}D}{\sqrt{5}G}\frac{1}{\sqrt{j}}$ and use the fact that
\begin{equation*} \label{eqn:anytime_fact}
	\begin{split}
		& \sum_{j=1}^t \frac{1}{\sqrt{j}} \leq \int_0^{t} \frac{1}{\sqrt{x}} \text{d} x=\left.2 \sqrt{x}\right|_0 ^{t}=2\sqrt{t}.
	\end{split}
\end{equation*}

\subsection{Proof of Theorem~\ref{thm:GDRO:col}}
The complete form of Theorem~\ref{thm:GDRO:col} is given below.
\begin{theorem} \label{thm:GDRO:col_formal}
    Let \textsc{Flex-GDRO} be Algorithm~\ref{alg:all_GDRO_two_layer} instantiated with Algorithm~\ref{alg:q_GDRO:col} as $\A_q$ and Algorithm~\ref{alg:w} as $\A_w$, with step sizes specified by Lemmas~\ref{lem:GDRO_q:col} and~\ref{lem:wt}. Under Assumptions~\ref{ass:gen_domain_w}--\ref{ass:gen_loss_convex}, for any fixed $t\in\N$ and any sample-size process $\{r_j\}_{j\in[t]}$ generated according to Definition~\ref{def:sample_size_protocol}, we have, with probability at least $1-\delta$,
	\begin{equation*}\label{eqn:cor:GDRO_all:high_formal}
		\begin{split}
            &\epsilon_{\phi}(\wb_t,\qb_t)\\
			\leq&\frac{DG}{\sqrt{t}}\left(2\sqrt{10}+8\sqrt{\ln\frac{4}{\delta}}\right)+\frac{1}{t}\sqrt{\sum_{j=1}^t\frac{m}{r_j}}\left(9\sqrt{2\ln m}+4\sqrt{2\ln\frac{20}{\delta}}\right)\\
            &+\frac{2^{1/4}m}{t}\sqrt{\ln m\sqrt{\sum_{j=1}^t\frac{m}{r_j}}\ln\frac{20m}{\delta}}+\frac{1}{t}\left(m+1+\frac{m^2\sqrt{\ln m}}{3}+\sqrt{\frac{2\sum_{j=1}^t\frac{m}{r_j}}{\ln m}}\right)\ln\frac{20}{\delta}\\
            &+\frac{1}{t}\left(\frac{m^2\sqrt{\ln m}}{3}+\frac{5m}{3}\right)\ln m+2\sqrt{\frac{2}{t}\ln\frac{4m}{\delta}}.
        \end{split}
	\end{equation*}
\end{theorem}

By substituting the results of Lemmas~\ref{lem:wt} and~\ref{lem:GDRO_q:col} into Lemma~\ref{lem:General_framework}, we complete the proof.

\subsection{Proof of Theorem~\ref{thm:GDRO:col_time_uniform}}
The full statement of Theorem~\ref{thm:GDRO:col_time_uniform} is given below.
\begin{theorem} \label{thm:GDRO:col_time_uniform_formal} 
Let \textsc{Flex-GDRO} be Algorithm~\ref{alg:all_GDRO_two_layer} instantiated with Algorithm~\ref{alg:q_GDRO:col} as $\A_q$ and Algorithm~\ref{alg:w} as $\A_w$, with step sizes specified by Lemmas~\ref{lem:GDRO_q:col} and~\ref{lem:wt}. 
Let $\{r_t\}_{t\ge 1}$ be any process generated according to Definition~\ref{def:sample_size_protocol}.
Under Assumptions~\ref{ass:gen_domain_w}--\ref{ass:gen_loss_convex}, with probability at least $1-\delta$, for all $t\in\N$,
\begin{equation*}
	\begin{split}
		\epsilon_{\phi}(\wb_t, \qb_t)
		\leq& \frac{DG}{\sqrt{t}}\left(2\sqrt{10}+10\sqrt{\ln\frac{16}{\delta}+2\ln\ln(2t)}\right) + 4\sqrt{\frac{1}{t}\left(\ln\frac{16m}{\delta}+2\ln\ln(2t)\right)}\\
		&+ \frac{1}{t}\sqrt{\sum_{j=1}^t\frac{m}{r_j}}\left(8\sqrt{\ln 	m}+2\sqrt{\ln\frac{20C_{2,t}}{\delta}}\right)+ \frac{m}{t}\sqrt{2\ln m\sqrt{\sum_{j\in M_t}\frac{m}{r_j}}\ln\left(\frac{20C_{1,t}}{\delta}\right)}\\
		&+\frac{2}{t}\sqrt{2\sum_{j\in M_t} \frac{m}{r_j}\ln \frac{20C_{3,t}}{\delta}}+\frac{m}{t}\ln\frac{20C_{2,t}}{\delta}+\frac{1}{t}\left(1+\sqrt{\frac{2\sum_{j=1}^t\frac{m}{r_j}}{\ln m}}\right)\ln \frac{80(\ln2t)^2}{\delta} \\
		& +\frac{m^2\sqrt{\ln m}}{t}\ln\left(\frac{20C_{1,t}}{\delta}\right) + \frac{2m\ln m}{t}+ \frac{2m}{t}\ln \frac{20C_{3,t}}{\delta},
	\end{split}
\end{equation*}
where $C_{1,t}=2\left(1+\ln^+ \left(\frac{2}{m}\sqrt{\sum_{j\in M_t}\frac{1}{r_j}}\right)\right)^2$, $C_{2,t}=2\left(1+\ln^+ \left(\frac{4}{m^2}\left(\sum_{j\in M_t} \frac{m}{r_j}+|S_t|\right)\right)\right)^2$, $C_{3,t}=2\left(1+\ln^+ \left(\sum_{j\in M_t} \frac{2}{mr_j}\right)\right)^2$ with $\ln^+ x = \max\{\ln x, 0\}$ for $x>0$ and $\ln^+0=0$, $S_t=\{j\in[t]\mid r_j=1\}$ and $M_t=\{j\in[t]\mid r_j\geq2\}$.
\end{theorem}

The analysis proceeds by decomposing the time-uniform optimization error and then bounding regrets for both the $w$-player and the $q$-player.
\begin{lemma} \label{lem:General_framework_time_uniform} 
	Under Assumptions~\ref{ass:gen_domain_w}-\ref{ass:gen_loss_convex}, suppose that
	\begin{compactenum}
		\item The $w$-player is equipped with a non-oblivious OCO algorithm $\A_w$, and  with probability at least $1 - \delta$, for all $t\in\N$, the regret $\Regw$ is upper bounded by $U_w(t,\delta)$.
		\item The $q$-player is equipped with a non-oblivious PLA algorithm $\A_q$, and  with probability at least $1 - \delta$, for all $t\in\N$, the regret $\Regqp$ is upper bounded by $U_q(t,\delta)$.
	\end{compactenum}
	Then, with probability at least $1 - \delta$, for all $t \in \N$,
		\begin{equation*} \label{eqn:lem:General_framework:result:high-pro-strong}
			\epsilon_{\phi}(\wb_t, \qb_t)
			\leq \frac{1}{t}U_w\left(t,\frac{\delta}{4}\right) + \frac{1}{t}U_q\left(t,\frac{\delta}{4}\right) + 4\sqrt{\frac{1}{t}\left(\ln\frac{16m}{\delta}+2\ln\ln(2t)\right)}.
		\end{equation*}
\end{lemma}
\begin{lemma} \label{lem:GDRO_q:col_time_uniform} 
	Let $\eta_{q,t}$ and $\gamma_t$ be defined in \eqref{eqn:proof:q:stepsize:1} and~\eqref{eqn:proof:q:stepsize:sep:gammat} for Algorithm~\ref{alg:q_GDRO:col}. Let $\{r_t\}_{t\ge 1}$ be any process generated according to Definition~\ref{def:sample_size_protocol}. Under Assumption~\ref{ass:gen_loss_bounded}, we obtain with probability at least $1-\delta$, for all $t\in\N$,
	\begin{equation*}
		\begin{split}
		\Regqp\leq & \sqrt{\sum_{j=1}^t\frac{m}{r_j}}\left(8\sqrt{\ln 	m}+2\sqrt{\ln\frac{5C_{2,t}}{\delta}}\right)+ m\sqrt{2\ln m\sqrt{\sum_{j\in M_t}\frac{m}{r_j}}\ln\left(\frac{5C_{1,t}}{\delta}\right)}\\
		&+2\sqrt{2\sum_{j\in M_t} \frac{m}{r_j}\ln \frac{5C_{3,t}}{\delta}}+m\ln\frac{5C_{2,t}}{\delta}+\left(1+\sqrt{\frac{2\sum_{j=1}^t\frac{m}{r_j}}{\ln m}}\right)\ln \frac{20(\ln2t)^2}{\delta} \\
		& +m^2\sqrt{\ln m}\ln\left(\frac{5C_{1,t}}{\delta}\right) + 2m\ln m+ 2m\ln \frac{5C_{3,t}}{\delta}.
		\end{split}
	\end{equation*}
where $C_{1,t}=2\left(1+\ln^+ \left(\frac{2}{m}\sqrt{\sum_{j\in M_t}\frac{1}{r_j}}\right)\right)^2$, $C_{2,t}=2\left(1+\ln^+ \left(\frac{4}{m^2}\left(\sum_{j\in M_t} \frac{m}{r_j}+|S_t|\right)\right)\right)^2$, $C_{3,t}=2\left(1+\ln^+ \left(\sum_{j\in M_t} \frac{2}{mr_j}\right)\right)^2$ with $\ln^+ x = \max\{\ln x, 0\}$ for $x>0$ and $\ln^+0=0$, $S_t=\{j\in[t]\mid r_j=1\}$ and $M_t=\{j\in[t]\mid r_j\geq2\}$.
\end{lemma}
\begin{lemma} \label{lem:wt_time_uniform}
	Under Assumptions~\ref{ass:gen_domain_w}, \ref{ass:gen_gradient_bounded} and \ref{ass:gen_loss_convex}, and setting $\eta_{w,t}=\frac{\sqrt{2}D}{\sqrt{5}G}\frac{1}{\sqrt{t}}$ for Algorithm~\ref{alg:w}, we have with probability at least $1-\delta$, for all $t\in\N$,
	\begin{equation*} \label{eqn:lem:wt_time_uniform}
		\Regw \leq DG\sqrt{t}\left(2\sqrt{10}+10\sqrt{\ln\frac{4}{\delta}+2\ln\ln(2t)}\right).
	\end{equation*}
\end{lemma}
By substituting the results of Lemmas~\ref{lem:GDRO_q:col_time_uniform} and~\ref{lem:wt_time_uniform} into Lemma~\ref{lem:General_framework_time_uniform}, we complete the proof.

\subsection{Proof of Lemma~\ref{lem:General_framework}} \label{sec:Proof of Lemma~lem:General_framework}
By Jensen's inequality and the outputs $\wb_t = \frac{1}{t} \sum_{j=1}^t \w_j$ and $\qb_t = \frac{1}{t} \sum_{j=1}^t \q_j$, we have
\begin{equation} \label{eqn:decom:error}
	\begin{split}
		& \epsilon_{\phi}(\wb_t, \qb_t) = \max_{\q\in \Delta_{m}}  \phi(\wb_t,\q)- \min_{\w\in \W}  \phi(\w,\qb_t)\\
		\leq & \frac{1}{t} \left(\max_{\q\in \Delta_{m}}  \sum_{j=1}^t \phi(\w_j,\q)- \min_{\w\in \W}  \sum_{j=1}^t \phi(\w,\q_j) \right) 
        = \frac{1}{t}\Regq + \frac{1}{t}\Regw,
	\end{split}
\end{equation}
where $\Regq$ and $\Regw$ are defined in~\eqref{eqn:decom:1}.
For the first term $\Regq$, by the definition of $\phi(\w,\q)$ in~\eqref{eqn:GDRO:convex:concave} and the property of linear optimization over the simplex $\Delta_m$, we have
\begin{equation} \label{eqn:decom:error:q}
	\begin{split}
		&\Regq \\
        =&  \max_{\q \in \Delta_m} \sum_{j=1}^t \phi(\w_j,\q) - \sum_{j=1}^t \phi(\w_j,\q_j) 
		=\max_{\q \in \Delta_m}  \sum_{i=1}^m q_i \left(\sum_{j=1}^t R_i(\w_j) \right) - \sum_{j=1}^t \sum_{i=1}^m q_{j,i}R_i(\w_j)\\
		= & \sum_{j=1}^t R_{k^*}(\w_j) - \sum_{j=1}^t \sum_{i=1}^m q_{j,i}R_i(\w_j)
		= \sum_{j=1}^t \sum_{i=1}^m q_{j,i} s_{j,i} - \sum_{j=1}^t s_{j,k^*} \\
		=&\sum_{j=1}^t \inner{\q_j}{\s_j} - \sum_{j=1}^t \inner{\q_j}{\hatsb_j} + \sum_{j=1}^t \inner{\q_j}{\hatsb_j} - \sum_{j=1}^t \hats_{j,k^*} + \sum_{j=1}^t \hats_{j,k^*} - \sum_{j=1}^t s_{j,k^*}\\
		\leq&\underbrace{\sum_{j=1}^t \inner{\q_j}{\s_j} - \sum_{j=1}^t \inner{\q_j}{\hatsb_j}}_{\mathtt{term (e_1)}} + \underbrace{\sum_{j=1}^t \inner{\q_j}{\hatsb_j} - \min_{i\in[m]} \sum_{j=1}^t \hats_{j,i}}_{\Regqp} + \underbrace{  \sum_{j=1}^t \hats_{j,k^*} - \sum_{j=1}^t s_{j,k^*}}_{\mathtt{term (e_2)}},
	\end{split}
\end{equation}
where $k^*=\argmax_{k\in [m]} \sum_{j=1}^tR_k(\w_{j})$, $\hatsb_j \in \R^m$ is defined in \eqref{def:hats:main_part} and the vector $\s_j \in \R^m$ is defined as
\begin{equation} \label{eqn:def:sti}
	s_{j,i} = \E_{j-1}[\hats_{j,i}]= 1 - R_i(\w_j) = 1 - \E_{\z \sim \P_i}[\ell(\w_j;\z)] \overset{\eqref{eqn:ass:value_bound}}{\in}  [0,1], \ \forall i \in [m].
\end{equation}

For $\mathtt{term (e_1)}$, we utilize the Hoeffding-Azuma inequality \citep{cesa2006prediction} in Lemma~\ref{lem:azuma} to establish a high-probability bound.
Denoting $V_j=\inner{\q_j}{\s_j}-\inner{\q_j}{\hatsb_j}$, by \eqref{eqn:def:sti} and \eqref{def:hats:main_part}, we know $\E_{j-1}\left[V_j\right]=0$ and $|V_j|\leq 1$ for all $j\in[t]$. By Lemma~\ref{lem:azuma}, we have for any fixed $t\in\N$, with probability at least $1-\delta$,
\begin{equation} \label{eqn:gen_e1}
	\mathtt{term (e_1)}=\sum_{j=1}^t V_j \leq \sqrt{2t \ln \frac{1}{\delta}}.
\end{equation}

Similarly, for $\mathtt{term (e_2)}$, we can verify that for all $i\in[m]$, $V_{j,i}^{\prime}=\hats_{j,i} - s_{j,i}$ is a martingale difference sequence bounded by 1, i.e., $\E_{j-1}[V_{j,i}^{\prime}]=0$ and $|V_{j,i}^{\prime}|\leq 1$ for all $j\in[t]$. Thus, by Lemma~\ref{lem:azuma} and taking the union bound over $i\in[m]$, we have for any fixed $t\in\N$, with probability at least $1-\delta$,
\begin{equation} \label{eqn:gen_e2}
	\mathtt{term (e_2)}=\sum_{j=1}^t V_{j,k^*}^{\prime} \leq \sqrt{2t \ln \frac{m}{\delta}}.
\end{equation}

By substituting \eqref{eqn:gen_e1} and \eqref{eqn:gen_e2} into \eqref{eqn:decom:error:q}, and subsequently combining the result with \eqref{eqn:decom:error}, we obtain the high-probability bound stated in \eqref{eqn:lem:General_framework:result:high-pro-weak}.

\subsection{Proof of Lemma~\ref{lem:Bernstein_inequality}}\label{sec:proof_Bernstein_inequality}
We first recall the classical Bernstein inequality~\citep[Lemma A.8]{cesa2006prediction}. Let $\{X_j\}_{j\in[t]}$ be a bounded martingale difference sequence satisfying $\E_{j-1}[X_j]=0$, $|X_j|\leq b$, and $\sum_{j=1}^t\E_{j-1}[X_j^2]\leq V$, where $b$ and $V$ are positive values. Then, with probability at least $1-\delta$,
\[
\sum_{j=1}^t X_j\leq \sqrt{2V\ln\frac{1}{\delta}}+\frac{2b}{3}\ln\frac{1}{\delta}.
\]

We now prove the adaptive-budget version. Fix any $t\in\mathbb{N}_+$ and denote $A_j=\sum_{s=1}^j\frac{m}{r_s}$ for $j\in[t]$. In particular, $A_t=\sum_{j=1}^t\frac{m}{r_j}$ is the cumulative budget quantity appearing in our variance bound. Since the sample sizes $r_j$ may depend on the history, $A_t$ may also vary with the realized history, so we cannot directly use $\Psi(A_t)$ as the variance parameter in the classical Bernstein inequality.

We handle this issue by dividing the possible range of $A_t$ into multiple scales. For each scale $B_k$, we stop the martingale when the cumulative budget $A_j$ exceeds $B_k$. This gives a martingale whose cumulative conditional variance is bounded by $\Psi(B_k)$, so the classical Bernstein inequality can be applied at this scale. We then take a union bound over all scales, so that all these bounds hold simultaneously. Finally, after the realized value of $A_t$ is known, we choose a scale $B_k$ satisfying $A_t<B_k\leq2A_t$, which gives the desired bound in terms of the realized budget $A_t$.

Let $L=1+\lfloor\log_2m\rfloor\leq m$ and set $B_k=2^{k+1}t$ for $k=0,\ldots,L-1$. For each $B_k$, define $X_j^{(B_k)}=X_j\ind[A_j\leq B_k]$. Since $A_j$ is $\mathcal{F}_{j-1}$-measurable under Definition~\ref{def:sample_size_protocol}, we have $\E_{j-1}[X_j^{(B_k)}]=\ind[A_j\leq B_k]\E_{j-1}[X_j]=0$ and $|X_j^{(B_k)}|\leq b$. Moreover, by the assumption on $\Psi$,
\[
\sum_{j=1}^t\E_{j-1}\left[(X_j^{(B_k)})^2\right]=\sum_{j=1}^t\ind[A_j\leq B_k]\E_{j-1}[X_j^2]\leq\Psi(B_k).
\]
If $\Psi(B_k)=0$, then $X_j^{(B_k)}=0$ for every $j\in[t]$. Otherwise, applying the classical Bernstein inequality to $\{X_j^{(B_k)}\}_{j\in[t]}$ with failure probability $\delta/m$, we obtain
\[
\Pr\left[\sum_{j=1}^tX_j^{(B_k)}>\sqrt{2\Psi(B_k)\ln\frac{m}{\delta}}+\frac{2b}{3}\ln\frac{m}{\delta}\right]\leq\frac{\delta}{m}.
\]
Taking a union bound over $k=0,\ldots,L-1$ and using $L\leq m$, with probability at least $1-\delta$, the following inequality holds simultaneously for all $k=0,\ldots,L-1$:
\[
\sum_{j=1}^tX_j^{(B_k)}\leq\sqrt{2\Psi(B_k)\ln\frac{m}{\delta}}+\frac{2b}{3}\ln\frac{m}{\delta}.
\]

It remains to select the scale corresponding to the realized value of $A_t$. Since $t\leq A_t\leq mt$, there exists $k\in\{0,\ldots,L-1\}$ such that $2^kt\leq A_t<2^{k+1}t=B_k\leq2A_t$. For this $k$, we have $A_j\leq A_t<B_k$ for every $j\in[t]$, and hence $X_j^{(B_k)}=X_j$ for every $j\in[t]$. Therefore, on the common high-probability event above,
\[
\sum_{j=1}^tX_j\leq\sqrt{2\Psi(B_k)\ln\frac{m}{\delta}}+\frac{2b}{3}\ln\frac{m}{\delta}.
\]
Since $\Psi$ is nondecreasing and $B_k\leq2A_t$, we further have $\Psi(B_k)\leq\Psi(2A_t)$. Recalling $A_t=\sum_{j=1}^t\frac{m}{r_j}$, we conclude that, with probability at least $1-\delta$,
\[
\sum_{j=1}^tX_j\leq\sqrt{2\Psi\left(2\sum_{j=1}^t\frac{m}{r_j}\right)\ln\frac{m}{\delta}}+\frac{2b}{3}\ln\frac{m}{\delta},
\]
which proves~\eqref{eqn:adaptive_budget_Bernstein}.

\subsection{Proof of Lemma~\ref{lem:Bernstein_inequality_time_uniform}} \label{sec:proof_Bernstein_inequality_time_uniform}
We begin by presenting a concentration inequality \citep{3d1c033e-51cd-3f22-8e4e-611c9a4badb9, lee2025minimax}. The proof in this section follows an online blog \footnote{\url{https://harinboy.github.io/posts/FreedmansInequality/}}.
\begin{lemma}   \citep[Proposition 4.]{lee2025minimax}
Let $\left\{X_t\right\}_{t=1}^{\infty}$ be a martingale difference sequence with respect to a filtration $\left\{\mathcal{F}_t\right\}_{t=0}^{\infty}$. Suppose $X_t \leq 1$ holds almost surely for all $t \geq 1$. Let $V_t:=\mathbb{E}\left[X_t^2 \mid \mathcal{F}_{t-1}\right]$ for $t \geq 1$ and take $\lambda>0$ and $\delta \in(0,1]$ arbitrarily. Then, the following inequality holds with probability at least $1-\delta$, for all $n \in \mathbb{N}$:
\begin{equation} \label{proof:lem:Bernstein_inequality_time_uniform_1}
	\sum_{t=1}^n X_t \leq \frac{e^\lambda-1-\lambda}{\lambda} \sum_{t=1}^n V_t+\frac{1}{\lambda} \log \frac{1}{\delta}.
\end{equation}
\end{lemma}

Suppose $0<\lambda<3$. Then, by Taylor expansion, it holds that
\begin{equation*}
	g(\lambda)=\sum_{i=0}^{\infty} \frac{\lambda^i}{(i+2)!} \leq \sum_{i=0}^{\infty} \frac{\lambda^i}{2 \cdot 3^i} \leq \frac{1}{2\left(1-\frac{\lambda}{3}\right)}
\end{equation*}
where the first inequality holds by $n!=2 \cdot 3 \ldots n \geq 2 \cdot 3 \ldots 3=2 \cdot 3^{n-2}$ for $n \geq 2$ and the last inequality holds since $0<\lambda<3$. Then, it holds that $\frac{e^\lambda-1-\lambda}{\lambda} \leq \frac{\lambda}{2\left(1-\frac{\lambda}{3}\right)}$. 
Plugging in this bound to \eqref{proof:lem:Bernstein_inequality_time_uniform_1},  we obtain that for fixed $0<\lambda<3$ and $T_n = \sum_{t=1}^n \mathbb{E}[X_t^2|\mathcal{F}_{t-1}]$, it holds that
\begin{equation}\label{proof:lem:Bernstein_inequality_time_uniform_2}
	\mathbb{P}\left(\exists n \in \mathbb{N}: \sum_{t=1}^n X_t \geq \frac{\lambda T_n}{2\left(1-\frac{\lambda}{3}\right)}+\frac{1}{\lambda} \log \frac{1}{\delta}\right) \leq \delta
\end{equation}
Taking the union bound over the following values of $\lambda=\lambda_k$ for $k=0,1, \ldots$, we have, with probabilities $\delta_k=\frac{\delta}{2(1+k)^2}$,
\begin{equation*}
	\lambda_k=\frac{3 \sqrt{\log \frac{1}{\delta_k}}}{\sqrt{\log \frac{1}{\delta_k}}+3 e^{\frac{k}{2}}}.
\end{equation*}
Plugging in this value to $\lambda$, we have
\begin{equation*}
\frac{\lambda_k T_n}{2\left(1-\frac{\lambda_k}{3}\right)}+\frac{1}{\lambda_k} \log \frac{1}{\delta_k}=\frac{T_n \sqrt{\log \frac{1}{\delta_k}}}{2 e^{\frac{k}{2}}}+e^{\frac{k}{2}} \sqrt{\log \frac{1}{\delta_k}}+\frac{1}{3} \log \frac{1}{\delta_k}.
\end{equation*}
By inequality \eqref{proof:lem:Bernstein_inequality_time_uniform_2} and taking the union bound, the probability of $\sum_{t=1}^n X_t$ exceeding this value for any $k$, or equivalently exceeding the minimum of these values over $k$, is less than $\delta$, i.e.,
\begin{equation*}
	\mathbb{P}\left(\exists n \in \mathbb{N}: \sum_{i=1}^n X_i \geq \min _{k=0,1, \ldots}\left\{\left(\frac{T_n}{2 e^{\frac{k}{2}}}+e^{\frac{k}{2}}\right) \sqrt{\log \frac{1}{\delta_k}}+\frac{1}{3} \log \frac{1}{\delta_k}\right\}\right) \leq \delta.
\end{equation*}
The proof is completed by showing that the minimum is smaller than the desired value by choosing appropriate $k=k_n$ for all $n$. We separately deal with the cases $T_n \geq 1$ and $0 \leq T_n<1$.

\textbf{Case 1} $(T_n \geq 1)$:

Choose $k_n=\left\lfloor\log T_n\right\rfloor$. Then, it holds that
\begin{equation*}
	\sqrt{T_n / e} \leq e^{\frac{k_n}{2}} \leq \sqrt{T_n}.
\end{equation*}
It follows that
\begin{equation*}
	\frac{T_n}{2 e^{\frac{k_n}{2}}}+e^{\frac{k_n}{2}} \leq \frac{\sqrt{e T_n}}{2}+\sqrt{T_n}\leq 2 \sqrt{T_n}.
\end{equation*}
Therefore, it holds that
\begin{align*}
	&\min _{k=0,1, \ldots}\left\{\left(\frac{T_n}{2 e^{\frac{k}{2}}}+e^{\frac{k}{2}}\right) \sqrt{\log \frac{1}{\delta_k}}+\frac{1}{3} \log \frac{1}{\delta_k}\right\} 
	\leq \left(\frac{T_n}{2 e^{\frac{k_n}{2}}}+e^{\frac{k_n}{2}}\right) \sqrt{\log \frac{1}{\delta_{k_n}}}+\frac{1}{3} \log \frac{1}{\delta_{k_n}} \\
	\leq & 2 \sqrt{T_n \log \frac{1}{\delta_{k_n}}}+\frac{1}{3} \log \frac{1}{\delta_{k_n}} 
	\leq 2 \sqrt{T_n \log \frac{2\left(1+\log ^{+} T_n\right)^2}{\delta}}+2 \log \frac{2\left(1+\log ^{+} T_n\right)^2}{\delta},
\end{align*}
where the last inequality uses that $\frac{1}{\delta_{k_n}}=\frac{2\left(1+k_n\right)^2}{\delta} \leq \frac{2\left(1+\log ^{+} T_n\right)^2}{\delta}$ and $\frac{1}{3} \leq 2$.

\textbf{Case 2} $(0 \leq T_n<1)$:

In this case, we choose $k_n=0$. Since $T_n \leq 1, T_n \leq \sqrt{T_n}$ holds. Using that $\log \frac{1}{\delta_0} \geq \log 2$, it holds that $\sqrt{\log \frac{1}{\delta_0}} \leq \frac{1}{\sqrt{\log 2}} \log \frac{1}{\delta_0}$. Therefore, it holds that
\begin{align*}
	\left(\frac{T_n}{2 e^0}+e^0\right) \sqrt{\log \frac{1}{\delta_0}}+\frac{1}{3} \log \frac{1}{\delta_0} 
	\leq &\sqrt{\frac{T_n}{2} \log \frac{1}{\delta_0}}+\frac{1}{\sqrt{\log 2}} \log \frac{1}{\delta_0}+\frac{1}{3} \log \frac{1}{\delta_0} \\
	\leq &2 \sqrt{T_n \log \frac{1}{\delta_0}}+2 \log \frac{1}{\delta_0}.
\end{align*}
Note that in this case, $\frac{1}{\delta_0}=\frac{2}{\delta}=\frac{2\left(1+\log ^{+} T_n\right)^2}{\delta}$.
The proof is complete. 

\subsection{Proof of Lemma~\ref{lem:high-prob-martingale:Exp3-IX}} \label{sec:proof-concentration}
The proof follows the argument of Lemma~1 of \citet{Exp3-IX}.
In this proof, write $\E_{j-1}[\cdot]=\E[\cdot\mid\mathcal{F}_{j-1}]$ and, for the one-step calculation, $\E_{j-1}^{*}[\cdot]=\E[\cdot\mid\mathcal{F}_{j-1},\hat{\xi}_{j,1},\ldots,\hat{\xi}_{j,m}]$. Thus the loss vector, $\p_j$, $\gamma_j$, the weights and $\ind[j\in S]$ are fixed under the latter conditioning, and the conditional distribution of $i_j$ remains $\p_j$.
For $p_{j,i}>0$, the IX estimator $\tilde{\xi}_{j,i}$ satisfies
\begin{equation}\label{eq:concentration-1}
\begin{split}
	\tilde{\xi}_{j,i} = \frac{\hat{\xi}_{j,i}}{p_{j,i} + \gamma_j} \cdot \indicator{i_j = i}  
	\leq & \frac{\hat{\xi}_{j,i}}{p_{j,i} + \gamma_j \hat{\xi}_{j,i}} \cdot\indicator{i_j = i} 
	= \frac{1}{2\gamma_j} \frac{2\gamma_j \cdot \hat{\xi}_{j,i}/p_{j,i}}{1 + \gamma_j \cdot \hat{\xi}_{j,i}/p_{j,i}} \cdot \indicator{i_j = i} \\
	\leq & \frac{1}{\beta_j} \log\left(1 + \beta_j \bar{\xi}_{j,i}\right),
\end{split}
\end{equation}
where the first inequality is obtained by $\hat{\xi}_{j,i} \in [0,1]$ for all $i \in [m]$ and $j \in [t]$, the last step is due to the inequality $\frac{z}{1 + z/2} \leq \log(1+z)$ for $z \geq 0$ and the notations $\beta_j = 2\gamma_j$ and $\bar{\xi}_{j,i} = (\hat{\xi}_{j,i}/p_{j,i}) \cdot \indicator{i_j = i}$ for $p_{j,i}>0$ are introduced for simplicity. If $p_{j,i}=0$, define $\bar{\xi}_{j,i}=0$. Since $\Pr(i_j=i\mid\mathcal{F}_{j-1},\hat{\xi}_{j,1},\ldots,\hat{\xi}_{j,m})=0$ in this case, $\tilde{\xi}_{j,i}=0$, so the final inequality in~\eqref{eq:concentration-1} also holds. In both cases, $\E_{j-1}^{*}[\bar{\xi}_{j,i}]\leq\hat{\xi}_{j,i}$.

We define the notations $\tilde{\lambda}_j = \ind[j\in S]\cdot\sum_{i=1}^m \alpha_{j,i} \tilde{\xi}_{j,i}$ and $\lambda_j = \ind[j\in S]\cdot\sum_{i=1}^m \alpha_{j,i} \hat{\xi}_{j,i}$. Then, we conclude that $\E_{j-1}^{*}[\exp(\tilde{\lambda}_j)]\leq\exp(\lambda_j)$ for all $j\in[t]$. This is true for $j\in[t]\setminus S$ clearly. For $j\in S$, we have
\begin{equation} \label{eq:concentration-2}
	\begin{split}
		\E_{j-1}^{*}\left[\exp(\tilde{\lambda}_j)\right] = {}& \E_{j-1}^{*}\left[\exp\left(\sum_{i=1}^m \alpha_{j,i} \tilde{\xi}_{j,i}\right)\right]
		\overset{\eqref{eq:concentration-1}}{\leq}  \E_{j-1}^{*}\left[\exp\left(\sum_{i=1}^m  \frac{\alpha_{j,i}}{\beta_j} \log\Big(1 + \beta_j \bar{\xi}_{j,i}\Big)\right)\right] \\
		\leq {}& \E_{j-1}^{*}\left[\exp\left(\sum_{i=1}^m  \log\Big(1 + \alpha_{j,i} \bar{\xi}_{j,i}\Big)\right)\right]
		= \E_{j-1}^{*}\left[\Pi_{i=1}^m \big(1 + \alpha_{j,i} \bar{\xi}_{j,i}\big)\right]  \\
		=& \E_{j-1}^{*}\left[1 + \sum_{i=1}^m \alpha_{j,i} \bar{\xi}_{j,i}\right]
		\leq 1 + \sum_{i=1}^m \alpha_{j,i} \hat{\xi}_{j,i}
		\leq \exp\left(\sum_{i=1}^m \alpha_{j,i} \hat{\xi}_{j,i} \right)  = \exp(\lambda_j),
	\end{split}
\end{equation}
where the second inequality uses $\frac{\alpha_{j,i}}{\beta_j}\leq1$ and $x\log(1+y)\leq\log(1+xy)$ for $y\geq-1$ and $x\in[0,1]$. The remaining steps use $\bar{\xi}_{j,i}\bar{\xi}_{j,k}=0$ for $i\ne k$, $\E_{j-1}^{*}[\bar{\xi}_{j,i}]\leq\hat{\xi}_{j,i}$, and $1+z\leq e^z$.

Multiplying the conditional bound in~\eqref{eq:concentration-2} by $\exp(-\lambda_j)$ gives
\[
 \E_{j-1}\left[\exp(\tilde{\lambda}_j-\lambda_j)\right]
 =\E_{j-1}\left[\E_{j-1}^{*}\left[\exp(\tilde{\lambda}_j-\lambda_j)\right]\right]\leq1.
\]
Consequently, $Z_n=\exp(\sum_{j=1}^n(\tilde{\lambda}_j-\lambda_j))$ is a non-negative supermartingale with $Z_0=1$, i.e., $\E_{n-1}[Z_n]\leq Z_{n-1}$. In particular, $\E[Z_t]\leq1$. By Markov's inequality,
\[
\Pr\left[\sum_{j=1}^t (\tilde{\lambda}_j - \lambda_j) > \epsilon\right] \leq \E\left[\exp\left( \sum_{j=1}^t (\tilde{\lambda}_j - \lambda_j) \right)\right] \cdot \exp(-\epsilon) \leq \exp(-\epsilon)
\]
holds for any $\epsilon >0$. By setting $\exp(-\epsilon) = \delta$ and noting that $\sum_{j=1}^t (\tilde{\lambda}_j - \lambda_j)=\sum_{j\in S} (\tilde{\lambda}_j - \lambda_j)$, we complete the proof.

\subsection{Proof of Lemma~\ref{lem:Bernstein_inequality_extension}}
Let $H_j=Q_j\ind[j\in M_t]$ and $X_j=H_j-\E_{j-1}[H_j]$. Since $r_j$ is $\mathcal{F}_{j-1}$-measurable, so is $\ind[j\in M_t]=\ind[r_j\geq2]$. By construction, $\E_{j-1}[X_j]=0$, so $\{X_j\}_{j\in[t]}$ is a martingale difference sequence. Moreover, by~\eqref{eqn:proof:q:stepsize:1}, we have $\eta_{q,j}\leq\sqrt{\ln m}$. On $j\in M_t$, the multi-sample estimator satisfies $\tildes_{j,i}\leq m$, and Lemma~\ref{lem:sample_bound} gives $\E_{j-1}[Q_j]\leq m\eta_{q,j}/(2r_j)$. Therefore, $0\leq H_j\leq m^2\sqrt{\ln m}/2$ and $|X_j|\leq m^2\sqrt{\ln m}/2$, and
\[
\E_{j-1}[X_j^2]\leq\E_{j-1}[H_j^2]\leq\frac{m^2\sqrt{\ln m}}{2}\E_{j-1}[H_j]\leq\frac{m^3\sqrt{\ln m}}{4}\frac{\eta_{q,j}}{r_j}\ind[j\in M_t].
\]

For any fixed $B>0$, let $n_B=\max\left(\left\{j\leq t:\sum_{s=1}^j\frac{m}{r_s}\leq B\right\}\cup\{0\}\right)$. By~\eqref{eqn:proof:q:stepsize:1} and Lemma~\ref{lem:varying_stepsize} with $\gamma=\frac{1}{2}$,
\[
\begin{split}
\sum_{j\in M_t}\frac{\eta_{q,j}}{r_j}\ind\left[\sum_{s=1}^j\frac{m}{r_s}\leq B\right]
&\leq\frac{\sqrt{\ln m}}{m}\sum_{j=1}^{n_B}\frac{\frac{m}{r_j}}{\sqrt{\sum_{s=1}^j\frac{m}{r_s}}}\\
&\overset{\eqref{eqn:lem:varying_stepsize}}{\leq}\frac{2\sqrt{\ln m}}{m}\sqrt{\sum_{j=1}^{n_B}\frac{m}{r_j}}
\leq\frac{2}{m}\sqrt{B\ln m}.
\end{split}
\]
Consequently,
\[
\begin{split}
\sum_{j=1}^t\E_{j-1}\left[\left(X_j\ind\left[\sum_{s=1}^j\frac{m}{r_s}\leq B\right]\right)^2\right]
&\leq\frac{m^3\sqrt{\ln m}}{4}\sum_{j\in M_t}\frac{\eta_{q,j}}{r_j}\ind\left[\sum_{s=1}^j\frac{m}{r_s}\leq B\right]\\
&\leq\frac{m^3\sqrt{\ln m}}{4}\cdot\frac{2}{m}\sqrt{B\ln m}\\
&=\frac{m^2\ln m}{2}\sqrt{B}
=:\Psi(B).
\end{split}
\]
Applying~\eqref{eqn:adaptive_budget_Bernstein} with this choice of $\Psi$, we obtain, with probability at least $1-\delta$,
\begin{equation}\label{eqn:lem:Bernstein_inequality:1}
\sum_{j=1}^t\left(H_j-\E_{j-1}[H_j]\right)\leq 2^{1/4}m\sqrt{\ln m\sqrt{\sum_{j=1}^t\frac{m}{r_j}}\ln\frac{m}{\delta}}+\frac{m^2\sqrt{\ln m}}{3}\ln\frac{m}{\delta}.
\end{equation}
Finally, since $\ind[j\in M_t]$ is $\mathcal{F}_{j-1}$-measurable, $\sum_{j=1}^t\E_{j-1}[H_j]=\sum_{j\in M_t}\E_{j-1}[Q_j]\leq\frac{m}{2}\sum_{j\in M_t}\frac{\eta_{q,j}}{r_j}.$
Combining this inequality with~\eqref{eqn:lem:Bernstein_inequality:1} yields
\[
\sum_{j\in M_t}Q_j\leq 2^{1/4}m\sqrt{\ln m\sqrt{\sum_{j=1}^t\frac{m}{r_j}}\ln\frac{m}{\delta}}+\frac{m^2\sqrt{\ln m}}{3}\ln\frac{m}{\delta}+\frac{m}{2}\sum_{j\in M_t}\frac{\eta_{q,j}}{r_j},
\]
which proves~\eqref{eqn:lem:Bernstein_inequality_extension}.

\subsection{Proof of Lemma~\ref{lem:General_framework_time_uniform}}
We adopt the notation introduced in Appendix~\ref{sec:Proof of Lemma~lem:General_framework} and continue from \eqref{eqn:decom:error:q}. We utilize Lemma~\ref{lem:azuma_time_uniform} to get the time-uniform bound for $\mathtt{term}(e_1)$ and $\mathtt{term}(e_2)$. According to Hoeffding’s inequality \citep[Lemma A.1.]{cesa2006prediction} and the conditions $\E_{t-1}[V_t] = 0$ and $|V_t| \leq 1$, as well as $\E_{t-1}[V_{t,i}^{\prime}] = 0$ and $|V_{t,i}^{\prime}| \leq 1$ for all $t \in \N, i\in[m]$, it follows that $\mathbb{E}_{t-1}[e^{s V_t}] \leq \exp\left(s^2/2\right)$ and $\mathbb{E}_{t-1}[e^{s V_{t,i}^{\prime}}] \leq \exp\left(s^2/2\right)$ for any $s \in \mathbb{R}$ and for all $i\in[m]$. Therefore, the sequences $\{V_t\}_{t=1}^\infty$ and $\{V_{t,i}^{\prime}\}_{t=1}^\infty$ for all $i \in [m]$ are conditionally 1-sub-Gaussian.

By Lemma~\ref{lem:azuma_time_uniform}, with probability at least $1-\delta$, we have for all $t\in\N$,
\begin{equation}\label{eqn:gen_e1_uniform}
	\mathtt{term (e_1)}=\sum_{j=1}^t V_j \leq 2^{\frac34}\sqrt{t\ln\frac{7(\ln2t)^2}{2\delta}}\leq 2\sqrt{t\left(\ln\frac{4}{\delta}+2\ln\ln(2t)\right)}.
\end{equation}
By Lemma~\ref{lem:azuma_time_uniform} and taking the union bound over $i\in[m]$, with probability at least $1-\delta$, we have for all $t\in\N$,
\begin{equation}\label{eqn:gen_e2_uniform}
	\mathtt{term (e_2)}=\sum_{j=1}^t V_{j,k^*}^{\prime} \leq 2^{\frac34}\sqrt{t\ln\frac{7m(\ln2t)^2}{2\delta}}\leq 2\sqrt{t\left(\ln\frac{4m}{\delta}+2\ln\ln(2t)\right)}.
\end{equation}
Substituting \eqref{eqn:gen_e1_uniform} and \eqref{eqn:gen_e2_uniform} into \eqref{eqn:decom:error:q}, and combining \eqref{eqn:decom:error}, we obtain the high-probability time-uniform bound holding for all $t\in\N$,
\begin{equation*}
	\epsilon_{\phi}(\wb_t, \qb_t)
	\leq \frac{1}{t}U_w\left(t,\frac{\delta}{4}\right) + \frac{1}{t}U_q\left(t,\frac{\delta}{4}\right) + 4\sqrt{\frac{1}{t}\left(\ln\frac{16m}{\delta}+2\ln\ln(2t)\right)}.
\end{equation*}

\subsection{Proof of Lemma~\ref{lem:GDRO_q:col_time_uniform}}
We adopt the notation introduced in Appendix~\ref{sec:proof_GDRO_q} and continue from \eqref{eq:proof:q:tl:2}.

For $\mathtt{term (d)}$, we extend Lemma~\ref{lem:high-prob-martingale:Exp3-IX} to the time-uniform version as follows.
\begin{lemma} \label{lem:high-prob-martingale:Exp3-IX_time_uniform}
	Assume the sampling and measurability conditions of Lemma~\ref{lem:high-prob-martingale:Exp3-IX} hold for all $j\in\N$: $\{\gamma_j\}_{j\geq1}$ is a positive, non-increasing process, and the non-negative weights satisfy $\alpha_{j,i}\leq2\gamma_j$.
    Let $\{r_t\}_{t\ge1}$ be a process generated according to Definition~\ref{def:sample_size_protocol}, and define $S_t=\{j\in[t]\mid r_j=1\}$. Then, with probability at least $1 -\delta$, for all $t$, we have
	\begin{equation} \label{eqn:high-prob-margingale-time-uniform}
		\sum_{j\in S_t} \sum_{i=1}^m \alpha_{j,i} (\tilde{\xi}_{j,i} - \hat{\xi}_{j,i}) \leq \ln\frac{4(\ln2t)^2}{\delta}.
	\end{equation}
\end{lemma}
Similar to \eqref{eq:proof:q:A} and by Lemma~\ref{lem:high-prob-martingale:Exp3-IX_time_uniform}, we have, with probability at least $1 -\delta$, for all $t\in\N$,
\begin{equation}  \label{eq:proof:q:A_time_uniform}
	\begin{split}
		\sum_{j\in S_t}\left(\frac{\eta_{q,j}}{2}+\gamma_j\right)\sum_{i=1}^{m}\tildes_{j,i}
		\overset{\eqref{eqn:proof:q:stepsize:sep:gammat}}{=} &\sum_{j\in S_t}\eta_{q,j}\sum_{i=1}^{m}\tildes_{j,i}
		\overset{\eqref{eqn:high-prob-margingale-time-uniform}}{\leq} \sum_{j\in S_t}\eta_{q,j}\sum_{i=1}^{m}\hats_{j,i}+\ln\frac{4(\ln2t)^2}{\delta}\\
		\overset{\eqref{def:hats:main_part}}{\leq} & \sum_{j\in S_t}\eta_{q,j}m+\ln\frac{4(\ln2t)^2}{\delta}
		\overset{\eqref{eqn:proof:q:stepsize:1}}{=} \sum_{j\in S_t}\sqrt{\frac{m\ln m}{\sum_{s=1}^j\frac{1}{r_s}}}+\ln\frac{4(\ln2t)^2}{\delta}\\
		\leq &\sum_{j\in S_t} \sqrt{\frac{m\ln m}{|[j]\cap S_t|}}+\ln\frac{4(\ln2t)^2}{\delta}
		\leq  2\sqrt{|S_t|m\ln m}+\ln\frac{4(\ln2t)^2}{\delta}.
	\end{split}
\end{equation}

For $\mathtt{term (e)}$, we extend Lemma~\ref{lem:Bernstein_inequality_extension} to the time-uniform version as follows.
\begin{lemma}   \label{lem:Bernstein_inequality_extension_time_uniform}
    Let $\{r_t\}_{t\geq1}$ follow Definition~\ref{def:sample_size_protocol}, so $r_t$ is $\mathcal{F}_{t-1}$-measurable, and define $M_t=\{j\in[t]\mid r_j\geq2\}$. Let $\{\eta_t\}_{t\geq1}$ be positive and non-increasing, with $\eta_t$ and $\q_t\in\Delta_m$ being $\mathcal{F}_{t-1}$-measurable for every $t\in\N$. For each round with $r_t\geq2$, define
    \[
    \tilde{s}_{t,i}=\frac{\hat{s}_{t,i}}{q_{t,i}+(1-q_{t,i})\frac{r_t-1}{m-1}}\ind[i\in C_t],
    \qquad Q_t=\frac{\eta_t}{2}\sum_{i=1}^m q_{t,i}\tilde{s}_{t,i}^2,
    \]
    where $|\hat{s}_{t,i}|\leq1$ and the random set $C_t\subseteq[m]$ satisfies
    \[
    \Pr[i\in C_t\mid\mathcal{F}_{t-1}]=q_{t,i}+(1-q_{t,i})\frac{r_t-1}{m-1},\qquad i\in[m].
    \]
    Set $Q_t=0$ when $r_t=1$. Then, with probability at least $1-\delta$, for all $t\in\N$,
	\begin{equation*} \label{eqn:lem:Bernstein_inequality_extension_time_uniform}
		\sum_{j \in M_t} Q_j
		\leq \eta_1 m^2 \ln \left(\frac{C_t}{\delta}\right)+ m\sqrt{\eta_1 m\sum_{j\in M_t} \frac{\eta_{j}}{r_j} \ln \left(\frac{C_t}{\delta}\right)} + \frac{m}{2} \sum_{j \in M_t} \frac{\eta_{j}}{r_j},
	\end{equation*}
	where $C_t=2\left(1+\ln^+ \left(\frac{1}{\eta_1 m}\sum_{j\in M_t} \frac{\eta_{j}}{r_j}\right)\right)^2$.
\end{lemma}
We denote $Q_t=\frac{\eta_{q,t}}{2}\sum_{i=1}^{m}q_{t,i}(\tildes_{t,i})^{2}$ for $t\in \N$. By Lemma~\ref{lem:Bernstein_inequality_extension_time_uniform}, applied with $\eta_t=\eta_{q,t}$, and $\sqrt{\frac{\ln m}{m}}\leq\eta_{q,1}\leq\sqrt{\ln m}$, we have, with probability at least $1-\delta$, for all $t\in \N$,
\begin{equation} \label{eq:proof:q:B_time_uniform}
	\begin{split}
		\sum_{j\in M_t} Q_j
		\leq & m\sqrt{\sqrt{\ln m} m\sum_{j\in M_t} \frac{\eta_{q,j}}{r_j} \ln 	\left(\frac{C_{1,t}}{\delta}\right)} + m^2\sqrt{\ln m} \ln \left(\frac{C_{1,t}}{\delta}\right) + \frac{m}{2} \sum_{j \in M_t} \frac{\eta_{q,j}}{r_j} \\
		\leq &m\sqrt{2\ln m\sqrt{\sum_{j\in M_t}\frac{m}{r_j}}\ln\left(\frac{C_{1,t}}{\delta}\right)}+m^2\sqrt{\ln m}\ln\left(\frac{C_{1,t}}{\delta}\right)+\sqrt{\sum_{j\in M_t}\frac{m}{r_j}\ln m},
	\end{split}
\end{equation}
where the last inequality is obtained by \eqref{eq:proof:q:B_2} and $C_{1,t}=2\left(1+\ln^+ \left(\frac{2}{m}\sqrt{\sum_{j\in M_t}\frac{1}{r_j}}\right)\right)^2$.

For $\mathtt{term (f)}$, we define $V_t=\inner{\q_t}{\hatsb_t}-\hats_{t,c_t}$ when $r_t=1$ and $V_t=\inner{\q_t}{\hatsb_t}-\inner{\q_t}{\tildesb_t}$ when $r_t\geq2$. Since $r_t$ and $\q_t$ are $\mathcal{F}_{t-1}$-measurable, the same conditional-expectation calculation as for fixed-time term (f) gives $\E_{t-1}[V_t]=0$ for all $t\in\N$. By Lemma~\ref{lem:sample_bound} and \eqref{def:hats:main_part}, we know $|V_t|\leq \frac{m}{2}$ for all $t\in\N$. Let $X_t=\frac{2V_t}{m}$, we have $|X_t|\leq1$ for all $t\in\N$. Moreover, for $r_t=1$, we have
\begin{equation*}
	\E_{t-1}\left[X_t^2\right]
	=\frac{4}{m^2}\E_{t-1}\left[\left(\inner{\q_t}{\hatsb_t}- \hats_{t,c_t}\right)^2\right]\leq \frac{4}{m^2},
\end{equation*}
and for $r_t\geq2$, we have
\begin{align*}
	\E_{t-1}\left[X_t^2\right]
	=&\frac{4}{m^2}\E_{t-1}\left[\left(\inner{\q_t}{\hatsb_t}-\inner{\q_t}{\tildesb_t}\right)^2\right]
	=\frac{4}{m^2}\left(\E_{t-1}\left[\left(\inner{\q_t}{\tildesb_t}\right)^2\right]-\E_{t-1}\left[\left(\inner{\q_t}{\hatsb_t}\right)^2\right]\right)\\
	\leq &\frac{4}{m^2}\E_{t-1}\left[\left(\inner{\q_t}{\tildesb_t}\right)^2\right]
	\overset{\eqref{eqn:lem:inequality_bound}}{\leq}\frac{4}{mr_t}\E_{t-1}\left[\inner{\q_t}{\tildesb_t}\right]\leq \frac{4}{mr_t}.
\end{align*}
Then, by Lemma~\ref{lem:Bernstein_inequality_time_uniform}, we have, with probability at least $1-\delta$, for all $t\in\N$,
\begin{equation} \label{eq:proof:q:C_time_uniform}
	\begin{split}
		\sum_{j=1}^t V_j
		=& \frac{m}{2} \sum_{j=1}^t X_j
		\leq  m \ln \frac{C_{2,t}}{\delta} + m\sqrt{\left(\frac{4}{m^2}\left(\sum_{j\in M_t} \frac{m}{r_j}+|S_t|\right)\right) \ln \frac{C_{2,t}}{\delta}}\\
		\leq&\sqrt{4\sum_{j=1}^t \frac{m}{r_j}\ln\frac{C_{2,t}}{\delta}}+m\ln\frac{C_{2,t}}{\delta}.
	\end{split}
\end{equation}
where $C_{2,t}=2\left(1+\ln^+ \left(\frac{4}{m^2}\left(\sum_{j\in M_t} \frac{m}{r_j}+|S_t|\right)\right)\right)^2$.

For $\mathtt{term (g)}$, we extend the fixed-weight construction preceding \eqref{eq:proof:q:D} to all times using Ville's inequality. For $n\in\N$ and $i\in[m]$, define $D_{n,i}:=\sum_{j\in S_n}(\tildes_{j,i}-\hats_{j,i})$. For a fixed $B>0$, let $\lambda_B=\sqrt{\ln(m)/B}$. The proof of Lemma~\ref{lem:high-prob-martingale:Exp3-IX}, applied with the weights $\alpha_{j,\ell}^{(B,i)}=\lambda_B\ind\left[\sum_{s=1}^j\frac{m}{r_s}\leq B\right]\ind[\ell=i]$, shows that
\[ Z_{n,i}^{(B)}=\exp\left(\lambda_B\sum_{j\in S_n}\ind\left[\sum_{s=1}^j\frac{m}{r_s}\leq B\right](\tildes_{j,i}-\hats_{j,i})\right) \]
is a non-negative supermartingale with $Z_{0,i}^{(B)}=1$.

Set $B_k=m2^{k+1}$ for $k\geq0$. Applying Ville's inequality with failure probabilities $6\delta/[m\pi^2(k+1)^2]$ and taking a union bound over all $i,k$, whose failure probabilities sum to $\delta$, yields a single event of probability at least $1-\delta$ on which $Z_{t,i}^{(B_k)}\leq m\pi^2(k+1)^2/(6\delta)$ for all $t,i,k$. Whenever $\sum_{j=1}^t\frac{m}{r_j}\leq B_k$, all historical indicators equal one, and hence $D_{t,i}\leq\lambda_{B_k}^{-1}\ln\frac{m\pi^2(k+1)^2}{6\delta}$.

If $\sum_{j=1}^t\frac{m}{r_j}<m$, then $S_t=\emptyset$ and $D_{t,i}=0$. Otherwise choose $k=\lfloor\log_2\left(\sum_{j=1}^t\frac{1}{r_j}\right)\rfloor$, so $\sum_{j=1}^t\frac{m}{r_j}\leq B_k\leq2\sum_{j=1}^t\frac{m}{r_j}$. Since $\sum_{j=1}^t\frac{m}{r_j}\leq mt$, we have $\lambda_{B_k}^{-1}\leq\sqrt{2}/(2\gamma_t)$ and $\pi^2(k+1)^2/6\leq\pi^2[\ln(2t)]^2/[6(\ln2)^2]\leq4[\ln(2t)]^2$. Thus, on the same event, simultaneously for all $i\in[m]$ and $t\in\N$,
\begin{equation}\label{eq:proof:q:D_time_uniform} 
	\sum_{j\in S_t}(\tildes_{j,i}-\hats_{j,i}) \leq\frac{\sqrt{2}}{2\gamma_t}\ln\frac{4m(\ln2t)^2}{\delta}=\sqrt{2\sum_{j=1}^t\frac{m}{r_j}\ln m}+\sqrt{\frac{2\sum_{j=1}^t\frac{m}{r_j}}{\ln m}}\ln\frac{4(\ln2t)^2}{\delta}. 
\end{equation}

For $\mathtt{term (h)}$, we define for all $i\in[m]$, $X_{t,i}=\frac{\tildes_{t,i}-\hats_{t,i}}{m}$ for $r_t\geq2$ and $X_{t,i}=0$ for $r_t=1$. It holds that $\E_{t-1}[X_{t,i}]=0$, $|X_{t,i}|\leq 1$ for all $i\in[m]$ and $t\in\N$.
When $r_t\geq2$, we have 
\begin{align*}
	\E_{t-1}[X_{t,i}^2]&=\frac{\E_{t-1}[\tildes_{t,i}^2]-\E_{t-1}[\hats_{t,i}^2]}{m^2}\leq \frac{\E_{t-1}[\tildes_{t,i}^2]}{m^2}\\
    &\overset{\eqref{def:hats:main_part}}{\leq}\frac{1}{m^2\left(q_{t,i}+(1-q_{t,i})\frac{r_t-1}{m-1}\right)}\leq \frac{1}{m\left(r_t-1\right)}\leq \frac{2}{mr_t}.
\end{align*}
Then by Lemma~\ref{lem:Bernstein_inequality_time_uniform} with failure probability $\delta/m$ for each coordinate and taking the union bound over $i\in[m]$, we have, with probability at least $1-\delta$, for all $i\in[m]$ and $t\in\N$,
\begin{equation} \label{eq:proof:q:E_time_uniform}
	\begin{split}
		\sum_{j\in M_t} \left(\tildes_{j,i}- \hats_{j,i}\right)
		=&m\sum_{j=1}^{t}X_{j,i}
		\leq  2\sqrt{\sum_{j\in M_t} \frac{2m}{r_j} \ln \frac{mC_{3,t}}{\delta}}+ 2m\ln \frac{mC_{3,t}}{\delta}\\
		\leq & 2\sqrt{\sum_{j\in M_t} \frac{2m}{r_j}}\left(\sqrt{\ln m}+\sqrt{\ln \frac{C_{3,t}}{\delta}}\right)+ 2m\ln m+ 2m\ln \frac{C_{3,t}}{\delta}
	\end{split}
\end{equation}
where $C_{3,t}=2\left(1+\ln^+ \left(\sum_{j\in M_t} \frac{2}{mr_j}\right)\right)^2$.

Combining \eqref{eq:proof:q:A_time_uniform}-\eqref{eq:proof:q:E_time_uniform} with \eqref{eq:proof:q:tl:2}, allocating failure probability $\delta/5$ to each term, and taking the union bound, we have with probability at least $1-\delta$, for all $t\in\N$,
\begin{equation*}\label{eqn:lem:GDRO_q:High-pro:Es_time_uniform}
	\begin{split}
		&\Regqp \\
		\leq {}& \sqrt{\sum_{j=1}^t\frac{m}{r_j}}\left((1+\sqrt{2})\sqrt{\ln m}+ 2\sqrt{\ln\frac{5C_{2,t}}{\delta}}\right)+ 2\sqrt{|S_t|m\ln m} + 4\sqrt{\sum_{j\in M_t} \frac{m}{r_j}\ln m}  \\
		&+ m\sqrt{2\ln m\sqrt{\sum_{j\in M_t}\frac{m}{r_j}}\ln\left(\frac{5C_{1,t}}{\delta}\right)}+2\sqrt{2\sum_{j\in M_t} \frac{m}{r_j}\ln \frac{5C_{3,t}}{\delta}} 
		+m\ln\frac{5C_{2,t}}{\delta} \\
        &+\left(1+\sqrt{\frac{2\sum_{j=1}^t\frac{m}{r_j}}{\ln m}}\right)\ln \frac{20(\ln2t)^2}{\delta}  +m^2\sqrt{\ln m}\ln\left(\frac{5C_{1,t}}{\delta}\right) + 2m\ln m+ 2m\ln \frac{5C_{3,t}}{\delta}\\
		\overset{\eqref{eqn:proof:result_scale}}{\leq} & \sqrt{\sum_{j=1}^t\frac{m}{r_j}}\left(8\sqrt{\ln m}+2\sqrt{\ln\frac{5C_{2,t}}{\delta}}\right)+ m\sqrt{2\ln m\sqrt{\sum_{j\in M_t}\frac{m}{r_j}}\ln\left(\frac{5C_{1,t}}{\delta}\right)}\\
		&+2\sqrt{2\sum_{j\in M_t} \frac{m}{r_j}\ln \frac{5C_{3,t}}{\delta}}+m\ln\frac{5C_{2,t}}{\delta}+\left(1+\sqrt{\frac{2\sum_{j=1}^t\frac{m}{r_j}}{\ln m}}\right)\ln \frac{20(\ln2t)^2}{\delta} \\
		& +m^2\sqrt{\ln m}\ln\left(\frac{5C_{1,t}}{\delta}\right) + 2m\ln m+ 2m\ln \frac{5C_{3,t}}{\delta}.
	\end{split}
\end{equation*}

\subsection{Proof of Lemma~\ref{lem:wt_time_uniform}}
We adopt the notation introduced in Appendix~\ref{sec:proof_xt} and continue from \eqref{eqn:smd:6}. To establish a high-probability bound for all $t\in\N$, we use Lemma~\ref{lem:azuma_time_uniform}, which provides a time-uniform bound. By Hoeffding’s inequality, $\E_{t-1}[V_{t}] = 0$ and \eqref{eqn:smd:upper_bound_Vt}, we know $\mathbb{E}_{t-1}[e^{s V_t}] \leq \exp\left(16D^2G^2s^2\right)$ for any $s \in \mathbb{R}$. Thus, the sequence $\{V_t\}_{t=1}^\infty$ is conditionally $4\sqrt{2}DG$-sub-Gaussian.
By Lemma~\ref{lem:azuma_time_uniform}, with probability at least $1-\delta$, we have for all $t\in\N$,
\begin{equation}  \label{eqn:smd:9}
	\sum_{j=1}^t V_j \leq 2^{\frac34}4\sqrt{2}DG\sqrt{t\ln\frac{7(\ln2t)^2}{2\delta}}\leq 10DG\sqrt{t\left(\ln\frac{4}{\delta}+2\ln\ln(2t)\right)}.
\end{equation}
Combining \eqref{eqn:lem:proof:xt:decom:2}, \eqref{eqn:smd:6} and \eqref{eqn:smd:9}, we obtain that with probability at least $1-\delta$, for all $t\in\N$, 
\begin{equation*}
	\begin{split}
		\sum_{j=1}^t \phi(\w_j,\q_j) - \min_{\w\in\W} \sum_{j=1}^t \phi(\w,\q_j)
		\leq DG\sqrt{t}\left(2\sqrt{10}+10\sqrt{\ln\frac{4}{\delta}+2\ln\ln(2t)}\right).
	\end{split}
\end{equation*}

\subsection{Proof of Lemma~\ref{lem:high-prob-martingale:Exp3-IX_time_uniform}} \label{sec:proof-concentration-time-uniform}
Under Definition~\ref{def:sample_size_protocol}, $r_j$ and hence $\ind[r_j=1]$ are $\mathcal{F}_{j-1}$-measurable for every $j\in\N$.
Following the notation and results in Section~\ref{sec:proof-concentration}, with $\ind[j\in S]$ replaced by $\ind[r_j=1]$, we know the process $Z_t = \exp(\sum_{j=1}^t (\tilde{\lambda}_j - \lambda_j))$ is a supermartingale. 
To obtain a time-uniform bound, we follow \citet{lee2025lasso} by partitioning $\N$ into intervals $I_k = \{t_k, t_k+1, \ldots, t_{k+1}-1\}$ where $t_k=2^k$ for $k\geq0$.
Applying Ville’s inequality over each $I_k$, we have for any $\delta>0$,
\begin{equation*}
	\mathbb{P}\left(\exists n\in I_k: Z_n\geq\frac{\pi^2\left(k+1\right)^2}{6\delta}\right)\leq \frac{6\delta}{\pi^2\left(k+1\right)^2}.
\end{equation*}
Equivalently, we obtain
\begin{equation*}
	\mathbb{P}\left(\exists n\in I_k: \sum_{j=1}^n (\tilde{\lambda}_j - \lambda_j)\geq\ln\frac{\pi^2\left(k+1\right)^2}{6\delta}\right)\leq \frac{6\delta}{\pi^2\left(k+1\right)^2}.
\end{equation*}

By $\frac{\pi^2(k+1)^2}{6}=\frac{\pi^2(\log_22t_k)^2}{6}=\frac{\pi^2}{6(\ln2)^2}(\ln2t_k)^2\leq4(\ln2n)^2,$ we get
\begin{equation}   \label{proof:Exp3-IX-time-uniform-res1}
	\mathbb{P}\left(\exists n\in I_k: \sum_{j=1}^n (\tilde{\lambda}_j - \lambda_j)\geq\ln\frac{4(\ln2n)^2}{\delta}\right)\leq \frac{6\delta}{\pi^2\left(k+1\right)^2}.
\end{equation}
Finally, by taking the union bound over $k \geq 0$ and using the fact $\sum_{k=0}^{\infty} \frac{1}{(k+1)^2} = \frac{\pi^2}{6}$, we complete the proof.

\subsection{Proof of Lemma~\ref{lem:Bernstein_inequality_extension_time_uniform}}
We define \(H_t = Q_t \cdot\ind[r_t\geq2]\) and $M_t=\{j|r_j\geq2,j\in[t]\}$. Since $r_t$ is $\mathcal{F}_{t-1}$-measurable, so is $\ind[r_t\geq2]$, and $\E_{t-1}[H_t]=\ind[r_t\geq2]\E_{t-1}[Q_t]$. 
First, we note that \(|\tilde{s}_{t,i}| \leq m\) for all $i \in [m]$ and $t\in\N$ with $r_t\geq2$. It follows that \(0 \leq Q_t \leq \frac{\eta_1}{2} m^2\), and that \(\mathbb{E}_{t-1}[Q_t] \leq \frac{m}{2} \frac{\eta_t}{r_t}\) for all \(t \in \N\) with $r_t\geq2$ by $|\hat{s}_{t,i}|\leq1$, the conditional inclusion probabilities, and Lemma~\ref{lem:sample_bound}. 
Let $X_t=\frac{2\left(H_t - \mathbb{E}_{t-1}[H_t]\right)}{\eta_1 m^2}$. Since $\eta_1>0$ is $\mathcal{F}_0$-measurable, $\E_{t-1}[X_t]=0$. It holds that $X_t \leq 1$ for all $t \in \N$, and its variance when $r_t\geq2$ is bounded as
\begin{align*}
	\E_{t-1}[X_t^2]&=\frac{4}{\eta^2_1 m^4}\E_{t-1}\left[\left(H_t-\E_{t-1}\left[H_t\right]\right)^2\right]\\
	&=\frac{4}{\eta^2_1 m^4}\left(\E_{t-1}\left[\left(H_t\right)^2\right]-\E_{t-1}\left[H_t\right]^2\right)
	\leq  \frac{4}{\eta^2_1 m^4}\E_{t-1}\left[\left(H_t\right)^2\right]
	\leq \frac{1}{\eta_1 m} \frac{\eta_{t}}{r_t}.
\end{align*}
When $r_t=1$, $H_t=X_t=0$. Thus the conditional variance sum is bounded by $\frac{1}{\eta_1m}\sum_{j\in M_t}\frac{\eta_j}{r_j}$. Applying Lemma~\ref{lem:Bernstein_inequality_time_uniform}, we conclude that with probability at least $1 - \delta$, for all $t\in\N$,
\begin{equation}   \label{eqn:lem:Bernstein_inequality:time-uniform}
	\begin{split}
		\sum_{j\in M_t}\left(Q_j-\E_{j-1}\left[Q_j\right]\right)
		=&\sum_{j=1}^t\left(H_j-\E_{j-1}\left[H_j\right]\right)=\frac{\eta_1 m^2}{2}\sum_{j=1}^tX_j\\
		\leq & \eta_1 m^2 \ln \left(\frac{C_t}{\delta}\right)+ m\sqrt{\eta_1 m\sum_{j\in M_t} \frac{\eta_{j}}{r_j} \ln \left(\frac{C_t}{\delta}\right)}.
	\end{split}
\end{equation}
where $C_t=2\left(1+\ln^+ \left(\frac{1}{\eta_1 m}\sum_{j\in M_t} \frac{\eta_{j}}{r_j}\right)\right)^2$.

Consequently, we have, with probability at least \(1-\delta\), for all $t\in\N$,
\begin{align*}
	\sum_{j \in M_t} Q_j 
    =& \sum_{j \in M_t} \left(Q_j - \mathbb{E}_{j-1}[Q_j]\right) + \sum_{j \in M_t} \mathbb{E}_{j-1}[Q_j] \\
	\overset{\eqref{eqn:lem:Bernstein_inequality:time-uniform}}{\leq}& \eta_1 m^2 \ln \left(\frac{C_t}{\delta}\right)+ m\sqrt{\eta_1 m\sum_{j \in M_t} \frac{\eta_{j}}{r_j} \ln \left(\frac{C_t}{\delta}\right)} + \frac{m}{2} \sum_{j \in M_t} \frac{\eta_{j}}{r_j}.
\end{align*}

\end{document}

%% file: tables/synthetic_diabetes.tex
\begin{tabular}{c c c c | c c c c}
\hline
\makecell{Risk \\ (Synthetic)} & \makecell{Time \\ (\textsc{Online}($1$))} & \cellcolor{gray!20}\makecell{Time \\ (\textsc{Flex-GDRO})} & Speedup & \makecell{Risk \\ (Diabetes)} & \makecell{Time \\ (\textsc{Online}($1$))} & \cellcolor{gray!20}\makecell{Time \\ (\textsc{Flex-GDRO})} & Speedup \\
\hline
0.525 & 98.192 & \cellcolor{gray!20}9.107 & 10.78 & 0.99 & 4.842 & \cellcolor{gray!20}0.569 & 8.50 \\
0.52 & 129.708 & \cellcolor{gray!20}12.129 & 10.69 & 0.985 & 6.479 & \cellcolor{gray!20}0.876 & 7.39 \\
0.515 & 175.201 & \cellcolor{gray!20}15.152 & 11.56 & 0.98 & 9.406 & \cellcolor{gray!20}1.236 & 7.61 \\
0.51 & 256.080 & \cellcolor{gray!20}21.191 & 12.08 & 0.975 & 13.495 & \cellcolor{gray!20}1.902 & 7.09 \\
0.505 & 432.405 & \cellcolor{gray!20}33.274 & 13.00 & 0.97 & 22.839 & \cellcolor{gray!20}3.488 & 6.55 \\
0.5 & 1123.537 & \cellcolor{gray!20}108.804 & 10.33 & 0.965 & 51.470 & \cellcolor{gray!20}8.710 & 5.91 \\
\hline
\end{tabular}

%% file: tables/civilcomments.tex
\begin{tabular}{c c c | c c c}
    \hline
    Risk & \makecell{Samples \\ (\textsc{Online}($1$))} & \cellcolor{gray!20}\makecell{Samples \\ (\textsc{Flex-GDRO})} & \makecell{Time \\ (\textsc{Online}($1$))} & \cellcolor{gray!20}\makecell{Time \\ (\textsc{Flex-GDRO})} & Speedup \\
    \hline
    0.68 & 61.35k & \cellcolor{gray!20}69.33k & 40.26 & \cellcolor{gray!20}4.79 & 8.40 \\
    0.67 & 67.55k & \cellcolor{gray!20}75.34k & 44.34 & \cellcolor{gray!20}5.20 & 8.52 \\
    0.66 & 77.75k & \cellcolor{gray!20}81.36k & 51.03 & \cellcolor{gray!20}5.62 & 9.08 \\
    0.65 & 97.81k & \cellcolor{gray!20}87.39k & 64.19 & \cellcolor{gray!20}6.03 & 10.64 \\
    0.64 & 120.00k & \cellcolor{gray!20}101.84k & 78.80 & \cellcolor{gray!20}7.03 & 11.20 \\
    \hline
    \end{tabular}

%% file: tables/femnist.tex
\begin{tabular}{c c c | c c c}
    \hline
    Risk & \makecell*{Samples (\(\times 10^3\)) \\ (\textsc{Online}($1$))} & \cellcolor{gray!20}\makecell*{Samples (\(\times 10^3\)) \\ (\textsc{Flex-GDRO})} & \makecell*{Time \\ (\textsc{Online}($1$))} & \cellcolor{gray!20}\makecell*{Time \\ (\textsc{Flex-GDRO})} & Speedup \\
    \hline
    0.98 & $52.06\pm0.65$ & \cellcolor{gray!20}$63.43\pm3.33$ & $78.45\pm0.90$ & \cellcolor{gray!20}$23.61\pm0.94$ & 3.32 \\
    0.95 & $71.22\pm1.19$ & \cellcolor{gray!20}$73.78\pm3.81$ & $106.50\pm1.69$ & \cellcolor{gray!20}$27.46\pm1.18$ & 3.88 \\
    0.94 & $79.53\pm0.30$ & \cellcolor{gray!20}$79.51\pm5.01$ & $118.76\pm0.54$ & \cellcolor{gray!20}$29.48\pm1.65$ & 4.03 \\
    0.93 & $89.73\pm1.68$ & \cellcolor{gray!20}$83.68\pm6.89$ & $133.66\pm2.70$ & \cellcolor{gray!20}$31.00\pm2.37$ & 4.31 \\
    \hline
\end{tabular}

%% file: tables/variance_weighted.tex
\begin{tabular}{rrrrrrr}
\toprule
$m/r$ & $m$ & $r$ & $T=1{,}000$ & $T=5{,}000$ & $T=10{,}000$ & $T=20{,}000$ \\
\midrule
5 & 50 & 10 & $2.476\pm0.133$ & $2.586\pm0.081$ & $2.680\pm0.092$ & $2.688\pm0.130$ \\
5 & 100 & 20 & $2.502\pm0.009$ & $2.669\pm0.038$ & $2.694\pm0.048$ & $2.703\pm0.073$ \\
5 & 200 & 40 & $2.525\pm0.081$ & $2.646\pm0.038$ & $2.691\pm0.082$ & $2.730\pm0.107$ \\
10 & 50 & 5 & $5.464\pm0.111$ & $5.687\pm0.160$ & $5.932\pm0.194$ & $6.060\pm0.225$ \\
10 & 100 & 10 & $5.538\pm0.159$ & $5.927\pm0.115$ & $5.925\pm0.064$ & $6.066\pm0.090$ \\
10 & 200 & 20 & $5.751\pm0.053$ & $5.862\pm0.147$ & $6.043\pm0.138$ & $6.044\pm0.214$ \\
25 & 50 & 2 & $14.242\pm0.462$ & $14.736\pm0.414$ & $15.469\pm0.230$ & $15.586\pm0.812$ \\
25 & 100 & 4 & $14.603\pm0.133$ & $15.463\pm0.354$ & $15.794\pm0.245$ & $15.739\pm0.477$ \\
25 & 200 & 8 & $15.038\pm0.315$ & $15.430\pm0.299$ & $15.730\pm0.389$ & $16.145\pm0.490$ \\
\bottomrule
\end{tabular}

%% file: tables/variance_unweighted.tex
\begin{tabular}{rrrrrrr}
\toprule
$m/r$ & $m$ & $r$ & $T=1{,}000$ & $T=5{,}000$ & $T=10{,}000$ & $T=20{,}000$ \\
\midrule
5 & 50 & 10 & $2.537\pm0.127$ & $2.635\pm0.068$ & $2.725\pm0.069$ & $2.739\pm0.136$ \\
5 & 100 & 20 & $2.549\pm0.010$ & $2.682\pm0.054$ & $2.719\pm0.049$ & $2.747\pm0.076$ \\
5 & 200 & 40 & $2.534\pm0.049$ & $2.667\pm0.041$ & $2.693\pm0.057$ & $2.753\pm0.109$ \\
10 & 50 & 5 & $5.597\pm0.092$ & $5.961\pm0.239$ & $6.146\pm0.174$ & $6.228\pm0.250$ \\
10 & 100 & 10 & $5.690\pm0.205$ & $6.099\pm0.140$ & $6.088\pm0.084$ & $6.181\pm0.153$ \\
10 & 200 & 20 & $5.801\pm0.054$ & $5.966\pm0.091$ & $6.134\pm0.141$ & $6.217\pm0.232$ \\
25 & 50 & 2 & $15.895\pm0.531$ & $16.461\pm0.626$ & $17.147\pm0.340$ & $17.189\pm0.888$ \\
25 & 100 & 4 & $15.296\pm0.342$ & $16.411\pm0.267$ & $16.531\pm0.290$ & $16.836\pm0.514$ \\
25 & 200 & 8 & $15.560\pm0.283$ & $15.985\pm0.259$ & $16.306\pm0.297$ & $16.605\pm0.567$ \\
\bottomrule
\end{tabular}

%% file: tables/variance_w_gradient.tex
\begin{tabular}{rrrrrrr}
\toprule
$m/r$ & $m$ & $r$ & $T=1{,}000$ & $T=5{,}000$ & $T=10{,}000$ & $T=20{,}000$ \\
\midrule
5 & 50 & 10 & $0.220\pm0.012$ & $0.189\pm0.009$ & $0.180\pm0.013$ & $0.183\pm0.011$ \\
5 & 100 & 20 & $0.212\pm0.003$ & $0.194\pm0.008$ & $0.179\pm0.006$ & $0.177\pm0.012$ \\
5 & 200 & 40 & $0.218\pm0.010$ & $0.192\pm0.007$ & $0.190\pm0.019$ & $0.183\pm0.007$ \\
10 & 50 & 5 & $0.213\pm0.004$ & $0.192\pm0.012$ & $0.180\pm0.010$ & $0.183\pm0.015$ \\
10 & 100 & 10 & $0.218\pm0.011$ & $0.191\pm0.012$ & $0.180\pm0.008$ & $0.172\pm0.012$ \\
10 & 200 & 20 & $0.213\pm0.005$ & $0.189\pm0.005$ & $0.180\pm0.011$ & $0.182\pm0.009$ \\
25 & 50 & 2 & $0.218\pm0.015$ & $0.192\pm0.018$ & $0.184\pm0.015$ & $0.182\pm0.009$ \\
25 & 100 & 4 & $0.218\pm0.002$ & $0.190\pm0.012$ & $0.179\pm0.005$ & $0.173\pm0.014$ \\
25 & 200 & 8 & $0.220\pm0.009$ & $0.191\pm0.008$ & $0.182\pm0.015$ & $0.180\pm0.010$ \\
\bottomrule
\end{tabular}